\documentclass[journal]{IEEEtran}
\usepackage{amsmath,amssymb,amsfonts}
\usepackage{cases}
\usepackage{color,xcolor}
\usepackage{graphicx}
\usepackage[caption=false,font=footnotesize]{subfig}
\usepackage{algorithm}
\usepackage{algorithmic}
\usepackage{epstopdf}
\usepackage{multirow}
\usepackage{rotating}
\usepackage{booktabs}
\usepackage{tikz}
\usepackage{acronym}
\usepackage{epstopdf}
\usepackage{pifont} % 导入pifont包以使用打钩和打叉符号

\usepackage[demo,
            export]{adjustbox}  % it also load graphicx
\usepackage{tabularray}
\usepackage{colortbl}
\usepackage[colorlinks,
linkcolor=blue,
anchorcolor=blue,
citecolor=blue
]{hyperref}
\usepackage{amssymb}
\usepackage{array} % 引入array宏包
\usepackage{catchfilebetweentags}
\definecolor{tableTitleColor}{rgb}{0.8,0.8,0.8}
\begin{document}

\title{Learning to Select Features with Manifold \\Deep Unfolding PCA}

\author{Jingjing Liu, Pei Tai, Xianchao Xiu, and Wanquan Liu, \IEEEmembership{Senior Member,~IEEE}

\thanks{This work was supported in part by the National Natural Science Foundation of China under Grant 62204044 and Grant 12371306, in part by the State Key Laboratory of Integrated Chips and Systems under Grant SKLICS-K202302, and in part by the National Key Research and Development Program of China under Grant 2025YFB3412002. (\textit{Corresponding author: Xianchao Xiu}.)}
\thanks{J. Liu and P. Tai are with the Shanghai Key Laboratory of Automobile Intelligent Network Interaction Chip and System,  School of Microelectronics,  Shanghai University, Shanghai 200444,  China (e-mail: \{jjliu, ptai\}@shu.edu.cn).}
\thanks{X. Xiu is with the School of Mechatronic Engineering and Automation,  Shanghai University,  Shanghai 200444, China (e-mail: xcxiu@shu.edu.cn).}
\thanks{W. Liu is with the School of Intelligent Systems Engineering, Sun Yat-sen University, Guangzhou 510275, China (e-mail: liuwq63@mail.sysu.edu.cn).}
}

\maketitle

\begin{abstract}
Unsupervised feature selection (UFS) is of great significance in image processing and pattern recognition. Nevertheless, existing model-based  methods often rely on manually tuned hyperparameters, while deep learning-based methods generally require massive training data. To address these issues, this paper proposes an efficient  manifold deep unfolding feature selection network (MDUFS-Net). Technically, we first incorporate the $\ell_{2,0}$-norm into the classic principal component analysis (PCA) framework, then unfold the alternating direction method of multipliers (ADMM)-based iterative algorithm into an end-to-end trainable network, thereby enabling automatic parameter learning. Moreover, a manifold optimization strategy is developed to improve the numerical robustness. On this basis, we further extend MDUFS-Net into an enhanced variant called MDUFS-Net+ by introducing the $\ell_{0}$-norm for more discriminative feature representations. Extensive experiments validate that our proposed MDUFS-Net and MDUFS-Net+ outperform state-of-the-art methods in the UFS task, with promising performance in clustering. Our code is available at \url{https://github.com/xianchaoxiu/MDUFS}.
\end{abstract}


\section{Introduction}

High-dimensional data have become increasingly prevalent in numerous practical applications such as machine learning, computer vision, and industrial engineering \cite{ray2021various}. However, such data often contain redundant features, which not only increase storage and computational costs but may also degrade the performance of subsequent learning tasks \cite{erhard2020advances}. Therefore, selecting an informative and representative feature subset from the original feature space without access to class labels has emerged as a critical issue in unsupervised feature selection (UFS); see, e.g., \cite{li2024exploring,jiao2023survey,li2017feature,theng2024feature} for recent surveys.

According to \cite{solorio2020review}, existing UFS methods can be generally categorized into three types: filter-based, wrapper-based, and embedded-based methods. Among them, filter-based methods evaluate feature importance by using statistical characteristics or structural information. One typical example is the Laplacian score (LapScore) \cite{he2005laplacian}, which offers the advantage of high computational efficiency. Embedded-based methods integrate feature selection with the learning process and demonstrate excellent performance. Representatives include unsupervised discriminative feature selection (UDFS) \cite{yang2011l21}, structured optimal graph feature selection (SOGFS) \cite{nie2016unsupervised}, robust neighborhood embedding (RNE) \cite{liu2020robust}, and others \cite{li2015unsupervised,zhou2024unsupervised,li2026unsupervised,shen2026causally}.

Recently, principal component analysis (PCA)-based UFS methods have received increasing attention. From a representation perspective, PCA aims to seek a low-dimensional subspace that preserves the dominant information \cite{greenacre2022principal}. However, conventional PCA usually constructs data representations through linear combinations of original features, which limits its interpretability to some extent \cite{xiu2023sparsity}. In order to alleviate this limitation, sparse PCA introduces sparsity constraints or regularization terms into PCA \cite{zou2006sparse}. For example, Li et al. \cite{li2023sparse} incorporated an $\ell_{2,p}$-norm regularization term with $p\in(0,1)$ into the PCA model and proposed sparse PCA feature selection (SPCAFS), which achieves feature sparsity while retaining the principal information. Nie et al. \cite{nie2023learning} proposed feature-sparsity constrained PCA (FSPCA) by imposing an $\ell_{2,0}$-norm constraint to enhance the row sparsity of the projection matrix, thereby selecting representative feature subsets. It should be noted that the aforementioned $\ell_{2,p}$-norm and $\ell_{2,0}$-norm are not norms in the strict sense, but we still refer to them as norms in this paper.
Recently, Gao et al. \cite{gao2024principal} combined the $\ell_{2,1}$-norm with the $\ell_{2,p}$-norm to better characterize the structure of the projection matrix, which is called FEN-PCAFS. Xiu et al. \cite{xiu2025bi} explored multiple sparsity patterns by integrating the $\ell_{2,p}$-norm with $\ell_{q}$-norm to develop bi-sparse UFS (BSUFS), where $p,q\in [0,1]$. Liu et al. \cite{liu2026bi} introduced an $\ell_{2,0}$-norm constraint and proposed a bi-level UFS method for hierarchical feature selection. Although these sparse modeling methods offer effective pathways for feature selection \cite{zhang2023structured,zhang2026top}, they suffer from a significant drawback: the involved regularization parameters in the mathematical models or the penalty parameters in the iterative algorithms should be determined manually rather than adjusted automatically, which limits their generalization capability.

Deep neural networks have been increasingly considered to UFS by exploiting representation learning to uncover latent structures from unlabeled data \cite{mirzaei2020deep,hassanieh2024selective,ling2025self}. However, these methods often lack an explicit connection with the mathematical models, making it difficult to maintain prior knowledge within the network. In contrast, deep unfolding, also known as deep unrolling, provides a model-driven learning framework to bridge this gap by mapping iterative optimization procedures into stage-wise trainable networks, thereby preserving model interpretability while enabling adaptive parameter learning \cite{monga2021algorithm,chen2022learning,shlezinger2023model}. In the past decades, deep unfolding has gained encouraging performance in various tasks, including compressed sensing \cite{guo2025usb}, image restoration \cite{liu2025star}, and wireless communication \cite{deka2026comprehensive}. However, to the best of our knowledge, it has not been investigated in the field of UFS. Furthermore, for PCA-based models with the $\ell_{2,0}$-norm, the induced non-convex property remains largely unexplored in deep unfolding. Another issue concerns the orthogonality constraint, which can be naturally formulated as an optimization problem on the stiefel manifold but requires high computational complexity \cite{boumal2020introduction,zhu2025sparse}. \textit{Therefore, how to incorporate the $\ell_{2,0}$-norm, adaptive parameter learning, and manifold preservation into a deep unfolding UFS framework remains a challenging problem.}

\begin{figure}[t]
\centering
\includegraphics[width=0.90\columnwidth]{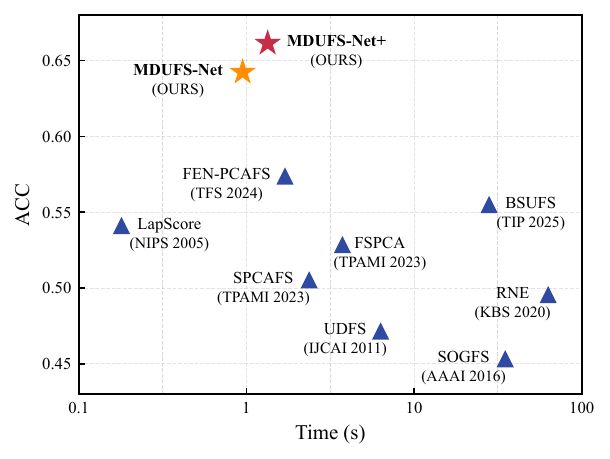}
\vspace{-0.3cm}
\caption{Performance-efficiency comparison of different methods based on  average accuracy (ACC) and runtime (s) across eight real-world datasets.}
\label{performance}
\end{figure}

Motivated by the above observations, this paper first constructs an $\ell_{2,0}$-norm regularized PCA-based UFS model and then develops an efficient deep unfolding framework according to the alternating direction method of multipliers (ADMM), thereby reducing the dependence on manual parameter tuning. Moreover, an effective manifold optimization technique is leveraged to preserve the orthogonality of the PCA projection matrix, resulting in the proposed MDUFS-Net. Building upon this, MDUFS-Net+ introduces an element-wise sparsity regularization term, which makes feature selection more robust. As demonstrated in Fig. \ref{performance}, compared with existing benchmark UFS methods, our proposed methods achieve high clustering performance with relatively low computational costs.

The main contributions of this paper can be summarized in the following three aspects:
\begin{itemize}
 \item We propose the first deep unfolding-based UFS method called MDUFS-Net, where the ADMM iterative algorithm is unfolded into a stage-wise trainable deep network, enabling end-to-end parameter learning.
\item We develop a manifold optimization mechanism into the stage-wise deep network updates, allowing the projection matrix to remain orthogonal throughout the unfolding process and enhancing network stability.
 \item We further extend MDUFS-Net into a general variant named MDUFS-Net+, where both row-wise and element-wise sparsity are enforced to improve feature selection performance in real-world scenarios.
\end{itemize}   

The structure of this paper is outlined as follows. 
Section \ref{preliminaries} introduces the related works. 
Section \ref{method} presents our proposed models and deep unfolding networks. Section \ref{experiment} provides the numerical results and ablation studies, and discussions. 
Section \ref{conclusion} concludes this paper and outlines future directions.

\section{Related Works}\label{preliminaries}

\subsection{Sparse PCA}

Given the data matrix $A\in \mathbb{R}^{d\times n}$, sparse PCA is formulated as the following mathematical model
\begin{equation}\label{SPCA}
\begin{aligned}
\min_{X}\quad & \frac{1}{2}\|A-XX^{\top}A\|_{\textrm{F}}^{2}
+\lambda \|X\|_{2,1} \\
\textrm{s.t.}~~~& X^{\top}X=I_{m},
\end{aligned}
\end{equation}
where $X\in \mathbb{R}^{d\times m}$ denotes the projection matrix, $\|X\|_{2,1}$ is the $\ell_{2,1}$-norm (defined as the sum of the $\ell_2$-norms of the rows), $\lambda>0$ is the parameter to determine the sparsity, and $X^\top X=I_m$ is the orthogonal constraint. In this paper, $m$ is omitted when there is no ambiguity. The first term in the objective is used for data reconstruction, while the second term induces row sparsity in the projection matrix for feature-level selection.

To obtain more discriminative feature representations, one can replace the $\ell_{2,1}$-norm in \eqref{SPCA} with $\ell_{2,p}$-norm, defined as $$\|X\|_{2,p}=\left(\sum_{i=1}^{m}\|\mathbf{x}_{i}\|^{p}\right)^{1/p},~p\in (0,1),$$
where $\mathbf{x}_{i}$ is the $i$-th row of $X$. Then, problem \eqref{SPCA}
leads to the following SPCAFS model
\begin{equation}\label{SPCAFS}
\begin{aligned}
\min_{X}\quad & \frac{1}{2}\|A-XX^{\top}A\|_{\textrm{F}}^{2}
+\lambda\|X\|^{p}_{2,p} \\
\textrm{s.t.}~~~& X^{\top}X=I.
\end{aligned}
\end{equation}
Note that the $\ell_{2,p}$-norm is non-convex, yet explicit solutions exist in special cases, such as $p=1/2$ \cite{xu20101,fan2026robust}. Moreover, $\ell_{2,p}$-norm facilitates obtaining sparser solutions than $\ell_{2,1}$-norm \cite{wan2026manifold}. Although SPCAFS achieves satisfactory feature selection results, the regularization parameter $\lambda$ must be carefully selected by cross-validation, and the corresponding optimization algorithm also involves tuning parameters, making the method difficult to implement on large-scale datasets.

\begin{figure*}[t]
\centering
\includegraphics[width=2\columnwidth]{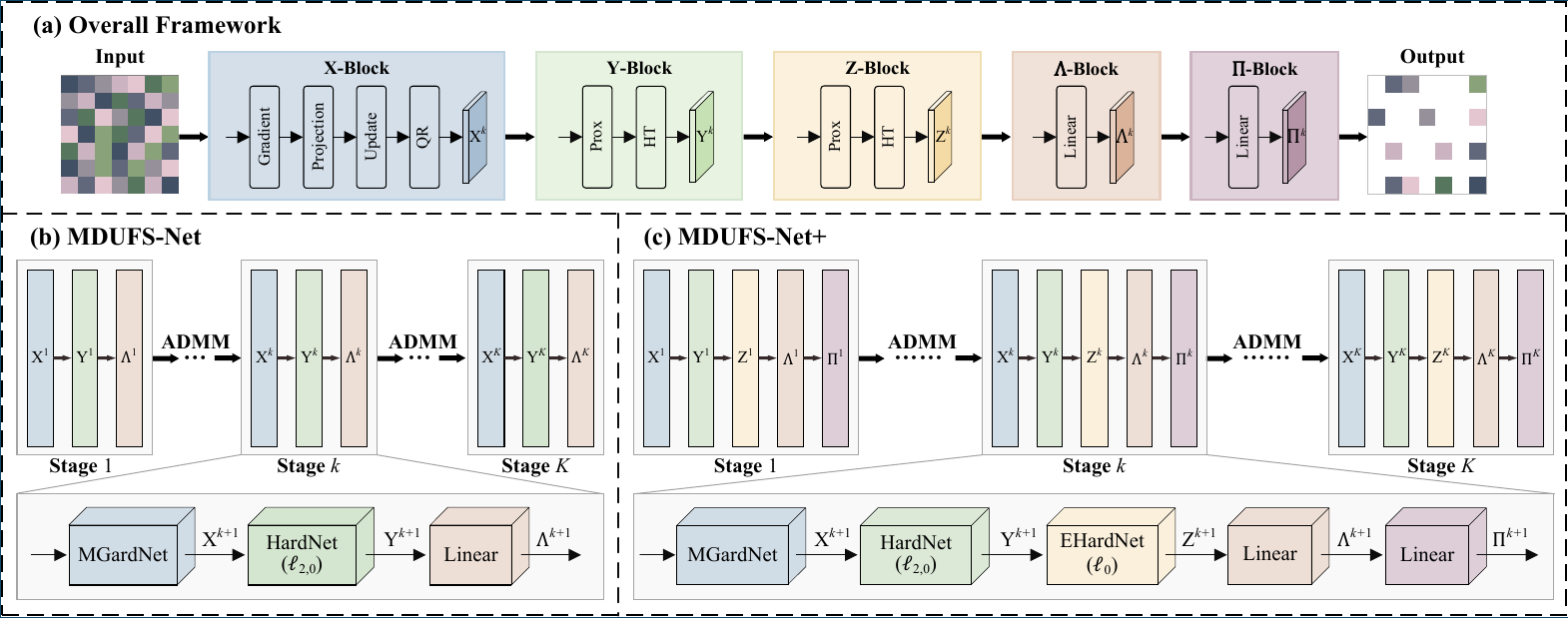}
\caption{Illustration of our proposed MDUFS-Net and MDUFS-Net+.}
\label{fig2}
\end{figure*}

\subsection{Deep Unfolding}

Deep unfolding has emerged as an effective strategy for solving sparse and structured optimization problems by designing trainable deep networks from iterative optimization procedures. Gregor and LeCun \cite{gregor2010learning} first unfolded the iterative process of sparse coding into a learnable network architecture. Yang et al. \cite{yang2016deep} further extended this idea by unfolding ADMM iterations and proposed deep ADMM-Net. Subsequent studies explored deep unfolding to various structured optimization models, including group-sparse nonnegative matrix factorization \cite{cui2023unrolling}, robust PCA \cite{joukovsky2024interpretable,wu2026rpcanet}, and low-rank sparse representation \cite{zhou2025low}. Interestingly, although deep unfolding has achieved significant success in image and signal processing, it has not been considered in the field of UFS, particularly in the context of PCA with non-convex sparsity regularization terms. In light of this, we attempt to introduce deep unfolding  with the aim of achieving superior feature selection performance.

\section{The Proposed Methods} \label{method}

This section presents the mathematical models and deep unfolding networks of MDUFS-Net and MDUFS-Net+, whose overall frameworks are given in Fig. \ref{fig2}.

\subsection{MDUFS-Net}
Given the excellent feature selection capability of the $\ell_{2,0}$-norm, this part considers the following model
\begin{equation}\label{raw_problem}
\begin{aligned}
\min_{X}\quad & \frac{1}{2}\|A-XX^{\top}A\|_{\textrm{F}}^{2}
+\lambda\|X\|_{2,0} \\
\textrm{s.t.}~~~& X^{\top}X=I,
\end{aligned}
\end{equation}
where $\|X\|_{2,0}$ is defined as the number of nonzero rows in $X$. As mentioned before, although problem \eqref{raw_problem} can be solved via iterative optimization, it is sensitive to parameter settings and lacks stability. Therefore, we explore below how to transform it into an end-to-end network using deep unfolding techniques.

We first define
\begin{equation}\label{equ:f}
f(X)=\frac12\|A-XX^\top A\|_{\textrm{F}}^{2},
\end{equation}
and introduce the auxiliary variable $Y=X$, then problem \eqref{raw_problem} is transformed to
\begin{equation}\label{5}
\begin{aligned}
\min_{X,Y}\quad & f(X)
+\lambda\|Y\|_{2,0} \\
\textrm{s.t.}~~~& X^{\top}X=I,~X=Y.
\end{aligned}
\end{equation}
The corresponding augmented Lagrangian function is
\begin{equation}\label{6}
    \begin{aligned}
        \mathcal{L}(X,Y,\Lambda)
        & = f(X)
        + \lambda\|Y\|_{2,0} \\
        & ~~~+\langle \Lambda, X-Y\rangle
        + \frac{\alpha}{2}\|X-Y\|_{\textrm{F}}^{2},
    \end{aligned}
\end{equation}
with $X^{\top}X=I$, where $\Lambda$ denotes the Lagrange multiplier and $\alpha > 0$ is the penalty parameter. Under the ADMM framework \cite{boyd2011distributed}, variables are updated via alternating minimization and unfolded into multiple network stages with learnable parameters. Specifically, the parameters  $\lambda, \alpha$ at the $k$-th stage are denoted as $\lambda_k, \alpha_k$, whereas in iterative optimization algorithms, these parameters are typically predefined.

\subsubsection{Update $X^{k+1}$ }
At the $k$-th stage, with $Y^k$ and $\Lambda^k$ fixed, the $X$-subproblem is written as
\begin{equation}\label{7}
    \begin{aligned}
        \min_X \quad
        & F_k(X)=f(X)
        + \frac{\alpha_k}{2}
        \|X-Y^k+\Lambda^k/\alpha_k\|_{\textrm{F}}^{2} \\
        \textrm{s.t.}~~~&X^{\top}X=I.
    \end{aligned}
\end{equation}
Since the orthogonality constraint $X^{\top}X=I$ prevents direct Euclidean updates from preserving feasibility, an effective manifold projection-based update strategy is employed during the deep unfolding process.

Let $G=AA^\top$. Then \eqref{equ:f} is expressed as
\begin{equation}\label{8}
    \begin{aligned}
        f(X)
        = \frac{1}{2}\|A\|_{\textrm{F}}^{2}
        - \frac{1}{2}\textrm{Tr}(X^{\top}GX).
    \end{aligned}
\end{equation}
Its Euclidean gradient is
\begin{equation}\label{9}
    \begin{aligned}
        \nabla f(X) = -GX,
    \end{aligned}
\end{equation}
and the gradient of the stage-wise objective function is 
\begin{equation}\label{10}
    \begin{aligned}
        \nabla F_k(X)
        = -GX
        + \alpha_k
        (
        X-Y^k+\Lambda^k/\alpha_k
        ).
    \end{aligned}
\end{equation}

To obtain an orthogonality-preserving update direction, the Euclidean gradient is projected onto a subspace-compatible tangent direction by
\begin{equation}\label{11}
    \begin{aligned}
        \mathcal{P}_X(Z)
        = (I-XX^{\top})Z,
    \end{aligned}
\end{equation}
where $Z$ denotes an arbitrary direction in the Euclidean space. Therefore, the constraint-preserving search direction is 

\begin{equation}\label{12}
    \begin{aligned}
        D_k
        &=
        \mathcal P_{X^k}
        (\nabla F_k(X^k))\\
        &=
        (I-X^k(X^k)^\top)
        \nabla F_k(X^k).
    \end{aligned}
\end{equation}

This projection enforces the orthogonality property of the search direction. Similar projection-based strategies have been explored for optimization problems with orthogonality constraints \cite{schechtman2023orthogonal}.
Subsequently, a stage-specific learnable step-size $\eta_k$ is introduced
\begin{equation}\label{13}
    \begin{aligned}
        \widetilde{X}^{k+1}
         =
         X^k-\eta_kD_k.
    \end{aligned}
\end{equation}
However, $\widetilde{X}^{k+1}$ may not exactly satisfy the orthogonality constraint. Let $\widetilde{X}^{k+1}=QR$, where $Q$ denotes the orthogonal factor satisfying $Q^\top Q=I$. Then
\begin{equation}\label{14}
   \begin{aligned}
      {X}^{k+1}=Q.
    \end{aligned}
\end{equation}
Overall, the solution of problem \eqref{7} is obtained by
\begin{equation}\label{15}
   \begin{aligned}
       X^{k+1}
       =
       \texttt{MGradNet}
       (
        X^k,Y^k,\Lambda^k,G,
        \alpha_k, \eta_k
       ),
    \end{aligned}
\end{equation}
where $\texttt{MGradNet}$ denotes the manifold gradient network. It embeds the orthogonality-preserving update into each unfolding stage, enabling optimization under the orthogonality constraint within the deep unfolding framework.

\subsubsection{Update $Y^{k+1}$}
Once $X^{k+1}$ has been updated, the $Y$-subproblem is given by
\begin{equation}\label{16}
    \begin{aligned}
        \min_Y\quad
        \lambda_k\|Y\|_{2,0}
        +
        \frac{\alpha_k}{2}
        \|X^{k+1}-Y+\Lambda^k/\alpha_k\|_{\textrm{F}}^{2},
    \end{aligned}
\end{equation}
which yields the closed-form solution
\begin{equation}\label{17}
    \begin{aligned}
        Y^{k+1}
        =
        \textrm{prox}_{\frac{\lambda_k}{\alpha_k}\|\cdot\|_{2,0}}
        (
        X^{k+1}
        +\Lambda^k/\alpha_k
        ),
    \end{aligned}
\end{equation}
see \cite{zhu2025joint} for more details.
For ease of description, denote
\begin{equation}\label{18}
    \begin{aligned}
         U^{k}
         =
        X^{k+1}
        +
        \Lambda^k/\alpha_k,
    \end{aligned}   
\end{equation}
then  for the $i$-th row vector $\mathbf{u}_i^{k}$, the row-wise update is
\begin{equation}\label{19}
    \begin{aligned}
        \mathbf y_i^{k+1}
                =
        \begin{cases}
                \mathbf{0}, & \|\mathbf{u}_i^{k}\|_2 \leq \tau_k,\\
                \mathbf{u}_i^{k}, & \|\mathbf u_i^{k}\|_2 > \tau_k,
        \end{cases}
    \end{aligned}
\end{equation}
where $\tau_k=\sqrt{2\lambda_k/\alpha_k}$ is the threshold to control the sparsity.
Therefore, the solution of problem \eqref{16} is characterized by 
\begin{equation}\label{20}
        Y^{k+1}
        =
        \texttt{HardNet}
        (U^{k},
        \tau_k),
\end{equation}
where $\texttt{HardNet}$ denotes the row-wise hard-thresholding operation at the $k$-th unfolding stage.

\begin{algorithm}[t]
    \caption{Deep unfolding for MDUFS-Net}
    \label{algorithm1}

    \textbf{Input:} Data $A$,
     stage-wise parameters $\lambda_k, \alpha_k, \eta_k$\\
    \textbf{Initialize:} $X^0, Y^0, \Lambda^0$\\
    \textbf{For} $k=0,1,\ldots,K-1$ \textbf{do}

    \begin{algorithmic}[1]
        \STATE Update $X^{k+1}$ according to
        \begin{equation*}
       X^{k+1}
       =
       \texttt{MGradNet}
       (
        X^k,Y^k,\Lambda^k,G,
        \alpha_k, \eta_k
       )
\end{equation*}

        \STATE Update $Y^{k+1}$ according to
        \begin{equation*}
        Y^{k+1}
        =
        \texttt{HardNet}
        (U^{k},
        \tau_k)
\end{equation*}

        \STATE Update $\Lambda^{k+1}$ according to
        \begin{equation*}
    \Lambda^{k+1}
    =
    \texttt{Linear}
    (\Lambda^k,X^{k+1},Y^{k+1},\alpha_k)
\end{equation*}
    \end{algorithmic}
    \textbf{End While}\\
    \textbf{Output:} $X^K, Y^K$
\end{algorithm}

\subsubsection{Update $\Lambda^{k+1}$}
When $X^{k+1}$ and $Y^{k+1}$ have been calculated, the Lagrange multiplier is updated according to 
\begin{equation}\label{21}
    \begin{aligned}
        \Lambda^{k+1}
        =
        \Lambda^k
        +
        \alpha_k
        (X^{k+1}-Y^{k+1}),
    \end{aligned}
\end{equation}
which can be solved by the following network
\begin{equation}\label{22}
    \Lambda^{k+1}
    =
    \texttt{Linear}
    (\Lambda^k,X^{k+1},Y^{k+1},\alpha_k),
\end{equation}
where $\texttt{Linear}(\cdot)$ represents the linear combination module for updating the dual variable at the $k$-th unfolding stage.

To sum up, our deep unfolding framework for MDUFS-Net is presented in Algorithm \ref{algorithm1}.

\subsection{MDUFS-Net+}

From the perspective of model-driven methods, the solution to problem \eqref{raw_problem} may still contain redundant elements. Therefore, this part introduces an additional $\ell_{0}$-norm regularization term, which is described by
\begin{equation}\label{23}
    \begin{aligned}
          \min_{X}\quad & \frac{1}{2}\|A-XX^{\top}A\|_{\textrm{F}}^{2}
          +\lambda\|X\|_{2,0}+\mu\|X\|_{0} \\
          \textrm{s.t.}~~~& X^{\top}X=I,
    \end{aligned}
\end{equation}
where $\|X\|_{0}$ is the $\ell_{0}$-norm, which counts the number of nonzero elements in $X$, and $\lambda, \mu>0$ are the parameters.

Recall the notation in \eqref{equ:f}, then the above problem \eqref{23} is equivalently restated as
\begin{equation}\label{24}
    \begin{aligned}
          \min_{X,Y,Z}\quad & f(X)
          +\lambda\|Y\|_{2,0}+\mu\|Z\|_{0} \\
          \textrm{s.t.}~~~~& X^{\top}X=I,~X=Y,~X=Z,
    \end{aligned}
\end{equation}
and the augmented Lagrangian function is
\begin{equation}\label{25}
    \begin{aligned}
        \mathcal{L}(X,Y,Z,\Lambda,\Pi)
        & = f(X)
        + \lambda\|Y\|_{2,0}+ \mu\|Z\|_{0} \\
        &~~~ +\langle \Lambda, X-Y\rangle
        + \frac{\alpha}{2} \|X-Y\|_{\textrm{F}}^{2}\\
        &~~~ + \langle \Pi, X - Z \rangle
        + \frac{\beta}{2} \|X - Z\|_{\textrm{F}}^{2},
    \end{aligned}
\end{equation}
where $\Lambda, \Pi$ are the corresponding Lagrange multipliers, and $\alpha,\beta\geq 0$ are the penalty parameters. It is observed that the update rule for variable $Y$ follows the same procedure as that in MDUFS-Net. Thus, its derivation is not repeated here.

\subsubsection{Update $X^{k+1}$}
At the $k$-th stage, the $X$-subproblem is  
\begin{equation}\label{26}
    \begin{aligned}
        \min_X \quad
        & \hat{F}_k(X)=f(X)
        + \frac{\alpha_k}{2}
        \|X-Y^k+\Lambda^k/\alpha_k\|_{\textrm{F}}^{2} \\
         &~~~~~~~~~~~~+ \frac{\beta_k}{2}
        \|X-Z^k+\Pi^k/\beta_k\|_{\textrm{F}}^{2} \\
        \textrm{s.t.}~~~& X^{\top}X=I.
    \end{aligned}
\end{equation}

Compared with MDUFS-Net, the above subproblem introduces an additional quadratic penalty term associated with the newly introduced variable  $Z^k$. Therefore, the Euclidean gradient is modified as
\begin{equation}\label{27}
    \begin{aligned}
        \nabla \hat{F}_k(X)
        &= -GX
        + \alpha_k(X-Y^k+\Lambda^k/\alpha_k)\\
        &~~~+ \beta_k(X-Z^k+\Pi^k/\beta_k),
    \end{aligned}
\end{equation}
and the projected search direction is 
\begin{equation}\label{28}
    \begin{aligned}
        D_k
        &=
        \mathcal P_{X^k}
        (\nabla \hat{F}_k(X^k))
        \\
        &=
        (I-X^k(X^k)^\top)
        \nabla \hat{F}_k(X^k).
    \end{aligned}
\end{equation}
The subsequent manifold gradient update with the learnable step-size $\eta_k$ and QR-based retraction follows the same procedure as MDUFS-Net. This extended update  is 
\begin{equation}\label{29}
\begin{aligned}
X^{k+1}
=
\texttt{MGradNet+}
(
X^k,Y^k,Z^k,
\Lambda^k,\Pi^k,
G,
\alpha_k,\beta_k,\eta_k
).
\end{aligned}
\end{equation}

\subsubsection{Update $Z^{k+1}$}
It is simplified to 
\begin{equation}\label{30}
        \min_Z\quad
        \mu_k\|Z\|_{0}
        +
        \frac{\beta_k}{2}
        \|X^{k+1}-Z+\Pi^k/\beta_k\|_{\textrm{F}}^{2}.
\end{equation}
According to the element-wise hard thresholding \cite{foucart2011hard}, problem \eqref{30} admits the following solution
\begin{equation}\label{31}
    \begin{aligned}
        Z^{k+1}
        =
        \textrm{prox}_{\frac{\mu_k}{\beta_k}\|\cdot\|_{0}}
        (
        X^{k+1}
        +\Pi^k/\beta_k
        ).
    \end{aligned}
\end{equation}
Specifically, denote
\begin{equation}\label{32}
    \begin{aligned}
         V^{k}
         =
        X^{k+1}
        +
        \Pi^k/\beta_k,
    \end{aligned}   
\end{equation}
and let $v_{i,j}^{k}$ be the $(i,j)$-th element of $V^k$. Then, the element-wise update is
\begin{equation}\label{33}
\begin{aligned}
z_{i,j}^{k+1}=
\begin{cases}
0, & | v_{i,j}^{k}|\leq\gamma_k,\\
v_{i,j}^{k}, & | v_{i,j}^{k}|>\gamma_k.
\end{cases}
\end{aligned}
\end{equation}
where $\gamma_k=\sqrt{2\mu_k/\beta_k}$.
In the deep unfolding, this operation is formulated as 
\begin{equation}\label{34}
    \begin{aligned}
        Z^{k+1}
        =
        \texttt{EHardNet}
        (V^{k},
        \gamma_k).
    \end{aligned}
\end{equation}

\begin{algorithm}[t]
    \caption{Deep unfolding for MDUFS-Net+}
    \label{algorithm2}

    \textbf{Input:} Data $A $,  stage-wise
    parameters $\lambda_k, \mu_k, \alpha_k, \beta_k, \eta_k$\\
    \textbf{Initialize:} $X^0, Y^0, Z^0, \Lambda^0, \Pi^0$\\
    \textbf{For} $k=0,1,\ldots,K-1$ \textbf{do}

    \begin{algorithmic}[1]

        \STATE Update $X^{k+1}$ according to \begin{equation*}
\begin{aligned}
X^{k+1}
=
\texttt{MGradNet+}
(
X^k,Y^k,Z^k,
\Lambda^k,\Pi^k,
G,
\alpha_k,\beta_k,\eta_k
)
\end{aligned}
\end{equation*}

        \STATE Update $Y^{k+1}$ according to \begin{equation*}
        Y^{k+1}
        =
        \texttt{HardNet}
        (U^{k},
        \tau_k)
\end{equation*}

        \STATE Update $Z^{k+1}$ according to \begin{equation*}
    \begin{aligned}
        Z^{k+1}
        =
        \texttt{EHardNet}
        (V^{k},
        \gamma_k)
    \end{aligned}
\end{equation*}

        \STATE Update $\Lambda^{k+1}$ according to 
        \begin{equation*}
    \Lambda^{k+1}
    =
    \texttt{Linear}
    (\Lambda^k,X^{k+1},Y^{k+1},\alpha_k)
\end{equation*}

        \STATE Update $\Pi^{k+1}$ according to \begin{equation*}
\Pi^{k+1}
=
\texttt{Linear}
(\Pi^k,X^{k+1},Z^{k+1},\beta_k)
\end{equation*}
    \end{algorithmic}
    \textbf{End While}\\
    \textbf{Output:} $X^K, Y^K, Z^K$
\end{algorithm}

\subsubsection{Update $\Lambda^{k+1}$ and $\Pi^{k+1}$}
The update of $\Lambda^{k+1}$ remains the same as that in MDUFS-Net, and 
$\Pi$ is updated via
\begin{equation}\label{35}
\Pi^{k+1}
=
\Pi^k
+
\beta_k
(X^{k+1}-Z^{k+1}),
\end{equation}
which is represented by 
\begin{equation}\label{36}
\Pi^{k+1}
=
\texttt{Linear}
(\Pi^k,X^{k+1},Z^{k+1},\beta_k).
\end{equation}

Finally, the deep unfolding scheme for our MDUFS-Net+ is summarized in Algorithm \ref{algorithm2}.

\subsection{Loss Function}

Given that the original mathematical models  involve the $\ell_{2,0}$-norm or $\ell_{0}$-norm, it cannot be directly optimized via gradient-based backpropagation. Therefore, we employ differentiable surrogate functions to approximate these regularization terms. Specifically, the differentiable surrogate function of the $\ell_{2,0}$-norm is defined as
\begin{equation}
\begin{aligned}
\varphi_{2,0}(Y)
= &
\sum_{i=1}^{d}
(1-\exp(-\|\mathbf{y}_{i}\|_2^2/(2\sigma^2))),
\end{aligned}
\end{equation}
where $\sigma>0$ is a smoothing parameter controlling the approximation of the surrogate function to the original norm. In our implementation, $\sigma$ is gradually adjusted during training.
Consequently, the loss function for  MDUFS-Net is
\begin{equation}\label{eq:MDUFS-Net_loss}
\begin{aligned}
\mathcal{L}_{\textrm{MDUFS-Net}}
=
f(X)
+
\lambda\varphi_{2,0}(Y).
\end{aligned}
\end{equation}

Similarly, the differentiable surrogate function of the $\ell_{0}$-norm is defined as
\begin{equation}
\begin{aligned}
\varphi_{0}(Z)
= &
\sum_{i=1}^{d}\sum_{j=1}^{m}
(1-\exp(-z_{ij}^{2}/(2\sigma^2))),
\end{aligned}
\end{equation}
and the loss function for MDUFS-Net+ is
\begin{equation}\label{eq:MDUFS-Net+_loss}
\begin{aligned}
\mathcal{L}_{\textrm{MDUFS-Net+}}
=
f(X)
+
\lambda\varphi_{2,0}(Y)
+
\mu\varphi_{0}(Z).
\end{aligned}
\end{equation}

\begin{table}[t]   \renewcommand\arraystretch{1.3}
    \caption{Statistical information of selected datasets.}\label{data}
    \centering
    \begin{tabular}{|c|c|c|c|c|}
        \hline
        Types & Datasets & Features & Samples & Classes \\
        \hline\hline
        \multirow{2}*{Synthetic}
         &Dartboard1 & 9 & 1000  & 4\\  
        \cline{2-5}
        &Diamond9  & 9& 3000 & 9 \\
        \hline
        \multirow{8}*{Real-world}
        &COIL20 & 1024 & 1440 & 20 \\
        \cline{2-5}
         &ISOLET & 617 & 1560 & 26 \\
        \cline{2-5}
        &USPS & 256 & 1000 & 10 \\
        \cline{2-5}
        &PIE & 1024 & 1166 & 53 \\
        \cline{2-5} 
        &LUNG & 3312 & 203 & 5 \\
        \cline{2-5}
        &MSTAR & 1024 & 2425  & 10 \\
        \cline{2-5}
        &MNIST & 784 & 10000 & 10 \\
        \cline{2-5}
        &HAR & 561 & 10299 & 6 \\
        \hline
    \end{tabular}
\end{table}

\begin{figure*}[!t]
\centering

\subfloat[LapScore]{%
\includegraphics[width=3.4cm]{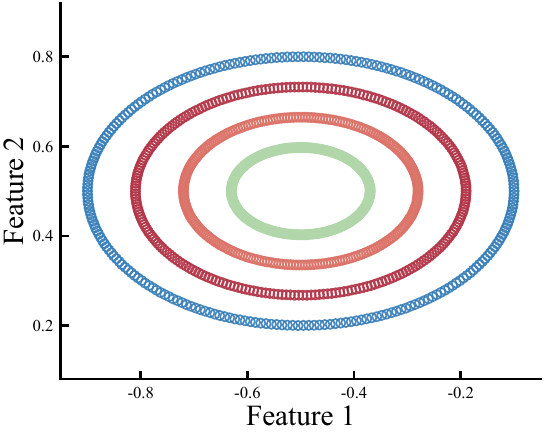}%
\label{fig:dartboard1_ls}%
}\hspace{-0.5mm}
\subfloat[SOGFS]{%
\includegraphics[width=3.4cm]{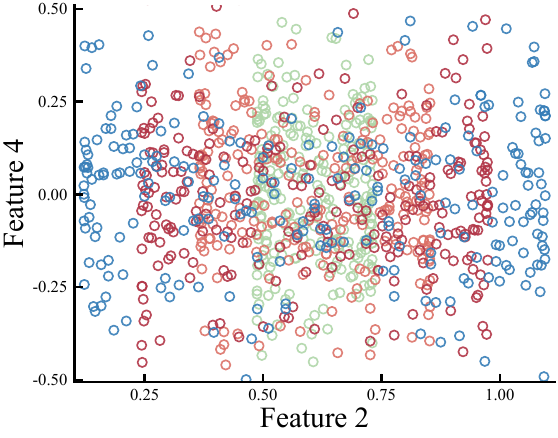}%
\label{fig:dartboard1_sogfs}%
}\hspace{-0.5mm}
\subfloat[RNE]{%
\includegraphics[width=3.4cm]{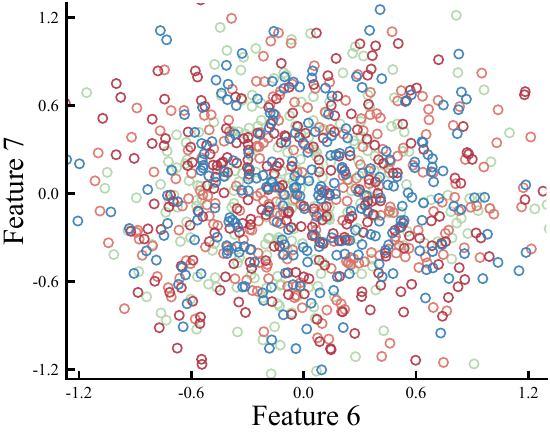}%
\label{fig:dartboard1_rne}%
}\hspace{-0.1mm}
\subfloat[UDFS]{%
\includegraphics[width=3.4cm]{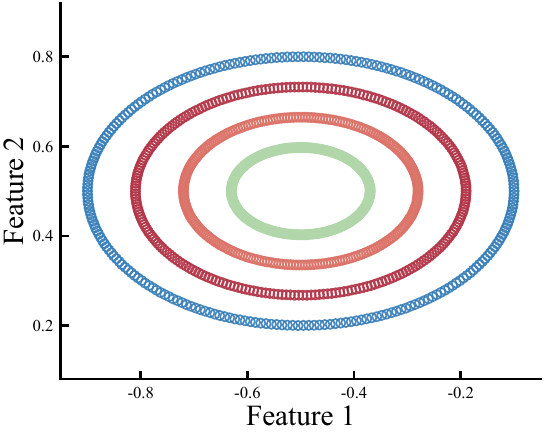}%
\label{fig:dartboard1_udfs}%
}\hspace{-0.5mm}
\subfloat[SPCAFS]{%
\includegraphics[width=3.4cm]{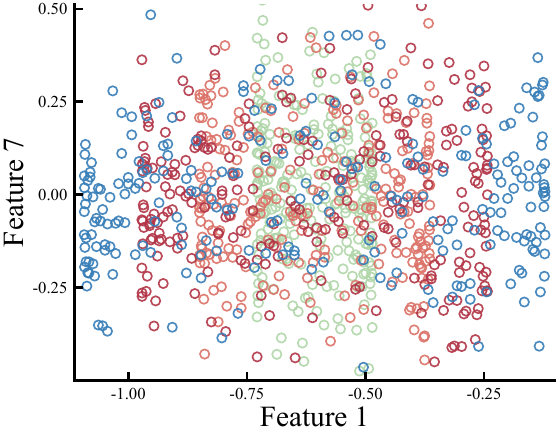}%
\label{fig:dartboard1_spcafs}%
}

\par\vspace{0.5mm}

\subfloat[FSPCA]{%
\includegraphics[width=3.4cm]{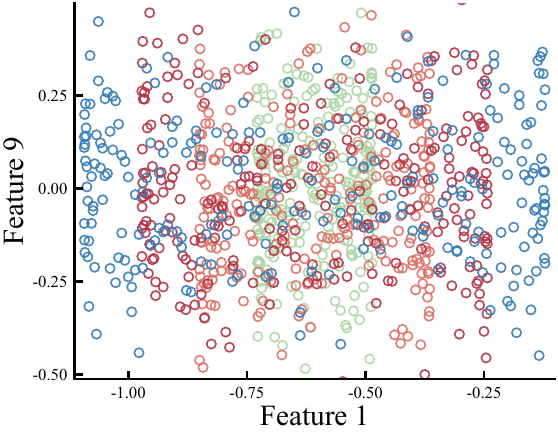}%
\label{fig:dartboard1_fspca}%
}\hspace{-0.5mm}
\subfloat[FEN-PCAFS]{%
\includegraphics[width=3.4cm]{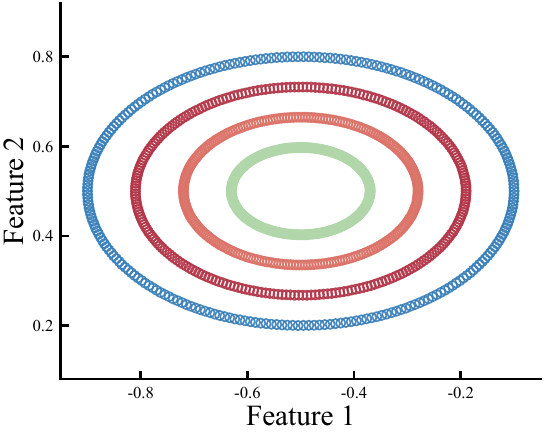}%
\label{fig:dartboard1_fenpcafs}%
}\hspace{-0.5mm}
\subfloat[BSUFS]{%
\includegraphics[width=3.4cm]{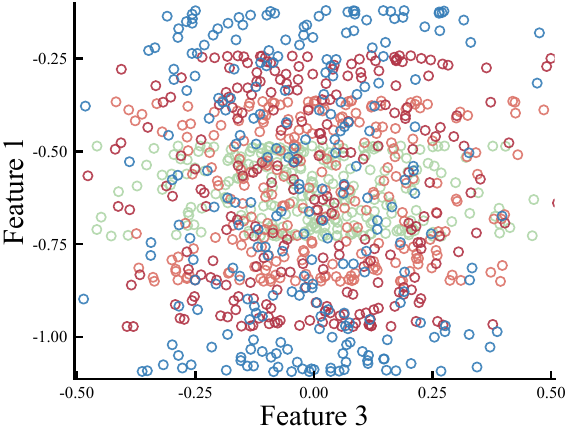}%
\label{fig:dartboard1_bsufs}%
}\hspace{-0.5mm}
\subfloat[MDUFS-Net]{%
\includegraphics[width=3.4cm]{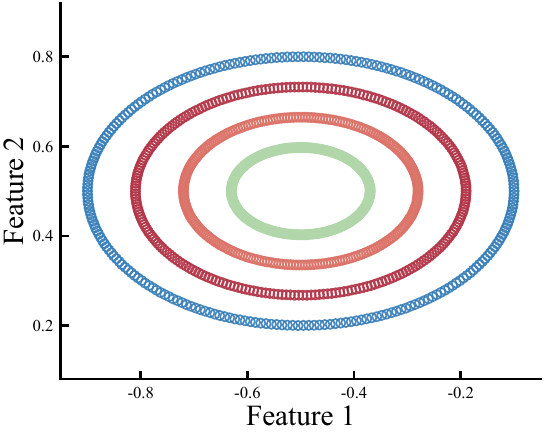}%
\label{fig:dartboard1_mdufs}%
}\hspace{-0.5mm}
\subfloat[MDUFS-Net+]{%
\includegraphics[width=3.4cm]{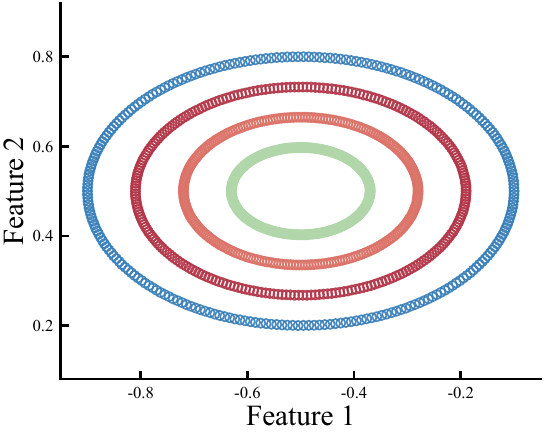}%
\label{fig:dartboard1_mdufsplus}%
}

\caption{Visual comparisons on the Dartboard1 dataset, where
(a)--(j) are the feature selection results.}
\label{dartboard1}

\end{figure*}

\begin{figure*}[!t]
\centering

\subfloat[LapScore]{%
\includegraphics[width=3.4cm]{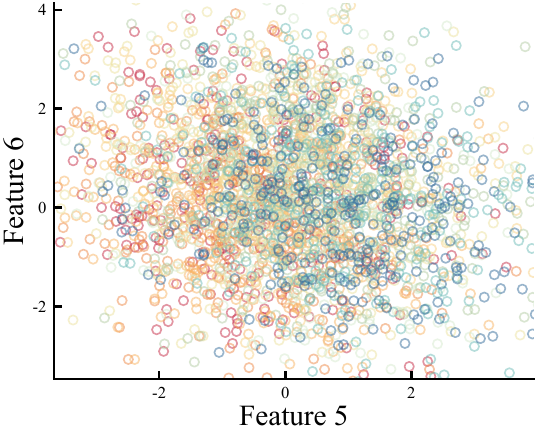}%
\label{fig:diamond9_ls}%
}\hspace{-0.5mm}
\subfloat[SOGFS]{%
\includegraphics[width=3.4cm]{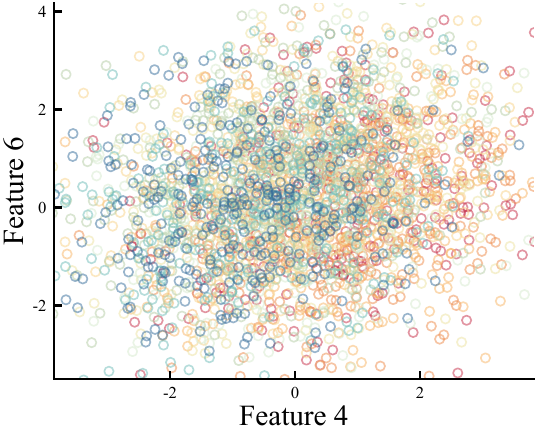}%
\label{fig:diamond9_sogfs}%
}\hspace{-0.5mm}
\subfloat[RNE]{%
\includegraphics[width=3.4cm]{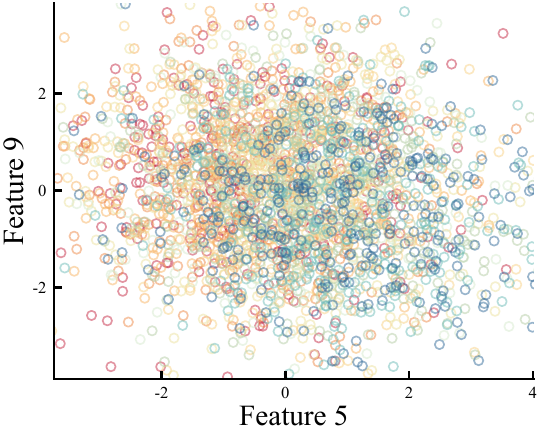}%
\label{fig:diamond9_rne}%
}\hspace{-0.5mm}
\subfloat[UDFS]{%
\includegraphics[width=3.4cm]{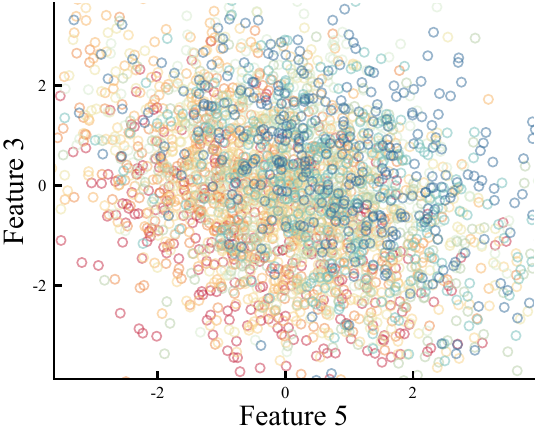}%
\label{fig:diamond9_udfs}%
}\hspace{-0.5mm}
\subfloat[SPCAFS]{%
\includegraphics[width=3.4cm]{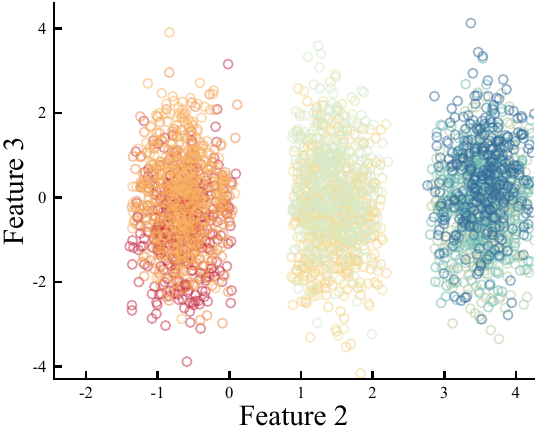}%
\label{fig:diamond9_spcafs}%
}

\par\vspace{0.5mm}

\subfloat[FSPCA]{%
\includegraphics[width=3.4cm]{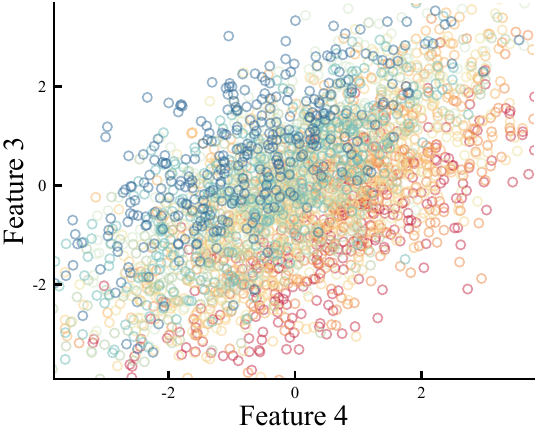}%
\label{fig:diamond9_fspca}%
}\hspace{-0.5mm}
\subfloat[FEN-PCAFS]{%
\includegraphics[width=3.4cm]{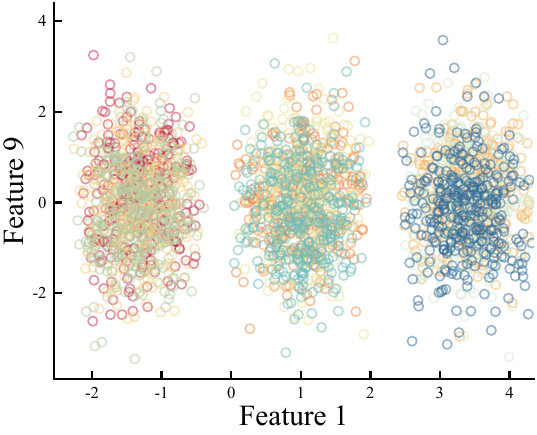}%
\label{fig:diamond9_fenpcafs}%
}\hspace{-0.5mm}
\subfloat[BSUFS]{%
\includegraphics[width=3.4cm]{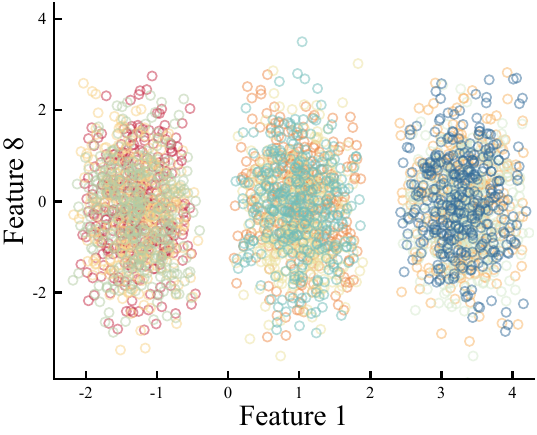}%
\label{fig:diamond9_bsufs}%
}\hspace{-0.5mm}
\subfloat[MDUFS-Net]{%
\includegraphics[width=3.4cm]{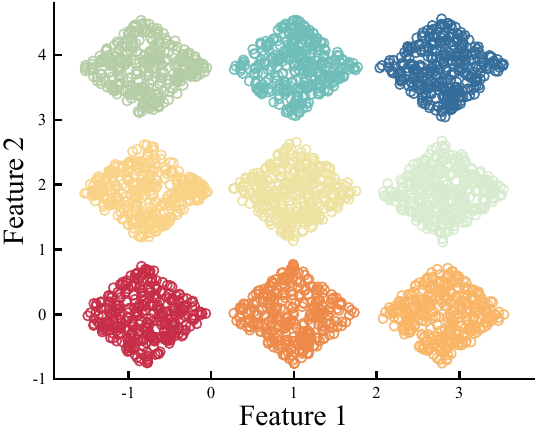}%
\label{fig:diamond9_mdufs}%
}\hspace{-0.5mm}
\subfloat[MDUFS-Net+]{%
\includegraphics[width=3.4cm]{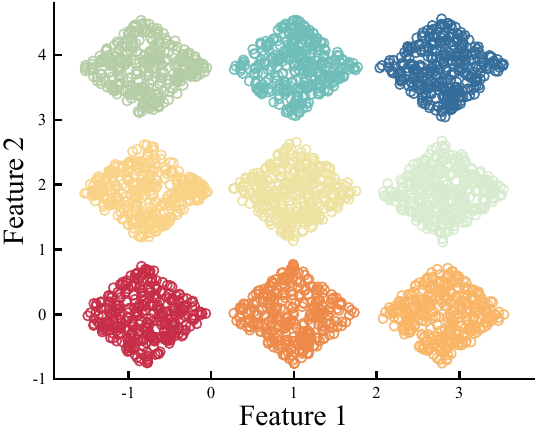}%
\label{fig:diamond9_mdufsplus}%
}

\caption{Visual comparisons on the Diamond9 dataset, where
(a)--(j) are the feature selection results.}
\label{Diamond9}

\end{figure*}
% -----------------Diamond9  end----------------------

% -----------------------ACC-----------------------
\begin{figure*}[t]
\centering

\subfloat[COIL20]{%
\includegraphics[width=4.3cm]{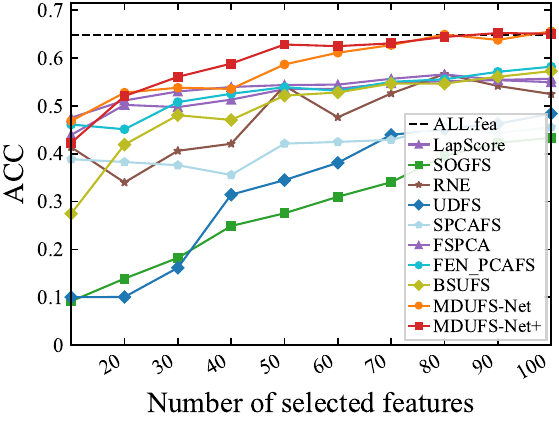}%
\label{fig:COIL20_ACC}%
}\hspace{1mm}
\subfloat[ISOLET]{%
\includegraphics[width=4.3cm]{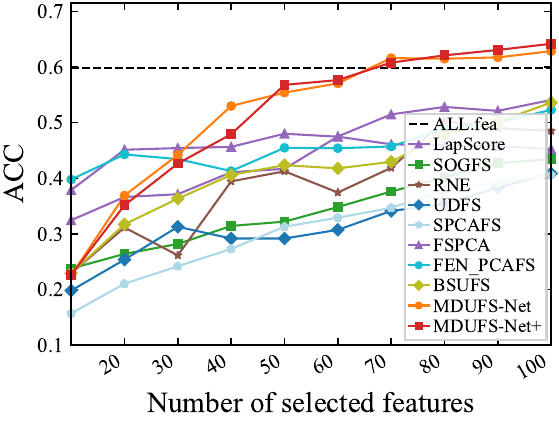}%
\label{fig:ISOLET_ACC}%
}\hspace{1mm}
\subfloat[USPS]{%
\includegraphics[width=4.3cm]{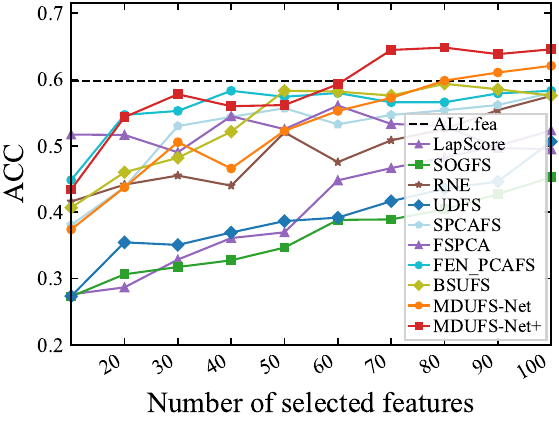}%
\label{fig:USPS_ACC}%
}\hspace{1mm}
\subfloat[PIE]{%
\includegraphics[width=4.3cm]{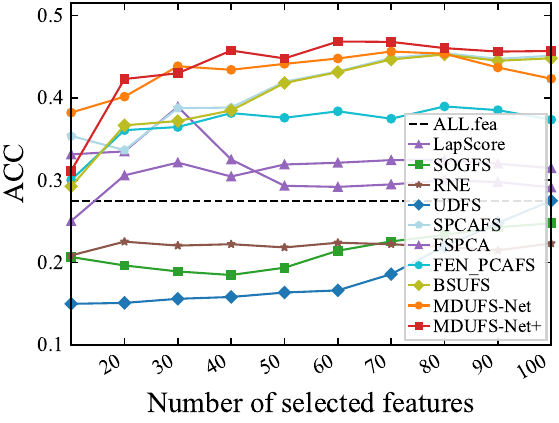}%
\label{fig:PIE_ACC}%
}

\par\vspace{0.5mm}

\subfloat[LUNG]{%
\includegraphics[width=4.3cm]{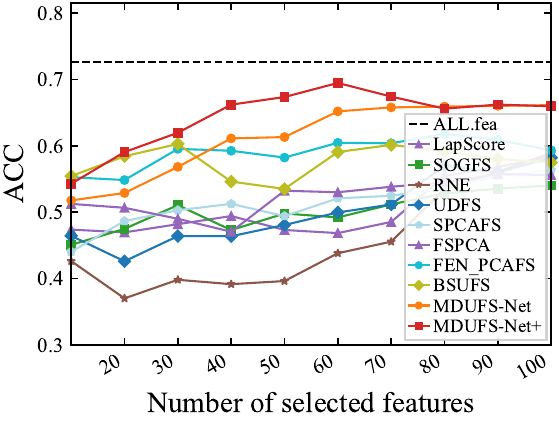}%
\label{fig:LUNG_ACC}%
}\hspace{1mm}
\subfloat[MSTAR]{%
\includegraphics[width=4.3cm]{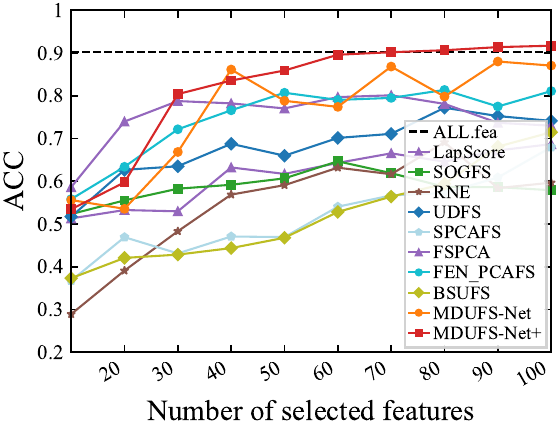}%
\label{fig:MSTAR_ACC}%
}\hspace{1mm}
\subfloat[MNIST]{%
\includegraphics[width=4.3cm]{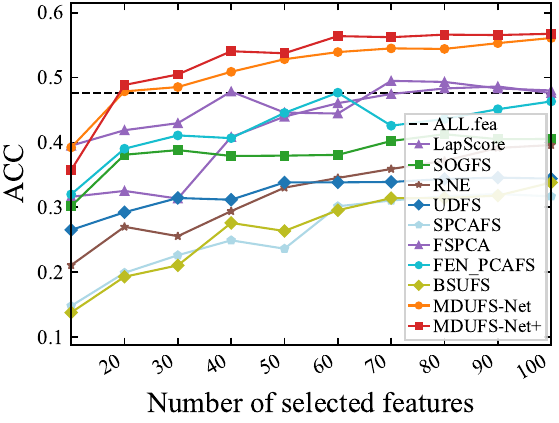}%
\label{fig:MNIST_ACC}%
}\hspace{1mm}
\subfloat[HAR]{%
\includegraphics[width=4.3cm]{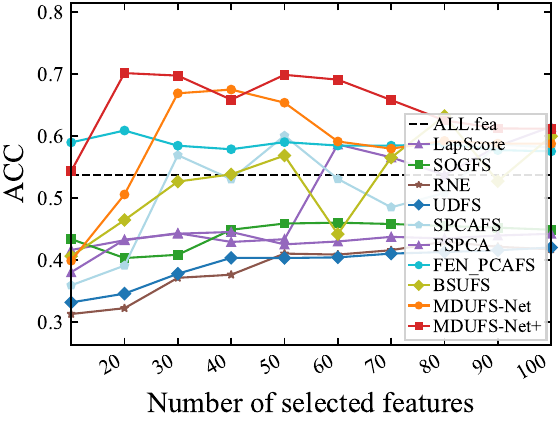}%
\label{fig:HAR_ACC}%
}

\caption{Visual comparisons of the ACC metric under different real-world datasets with different numbers of selected features.}
\label{ACC}

\end{figure*}
% ---------------------------NMI-----------------------
\begin{figure*}[t]
\centering

\subfloat[COIL20]{%
\includegraphics[width=4.3cm]{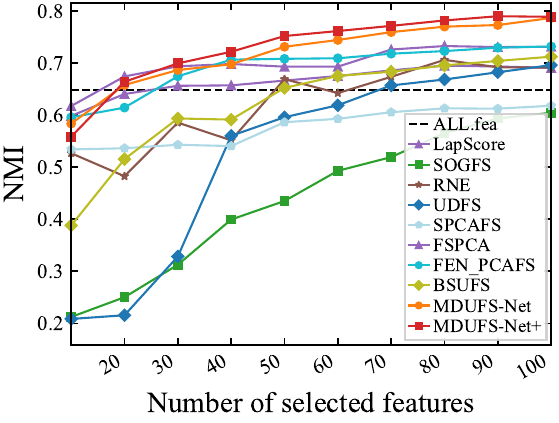}%
\label{fig:COIL20_NMI}%
}\hspace{1mm}
\subfloat[ISOLET]{%
\includegraphics[width=4.3cm]{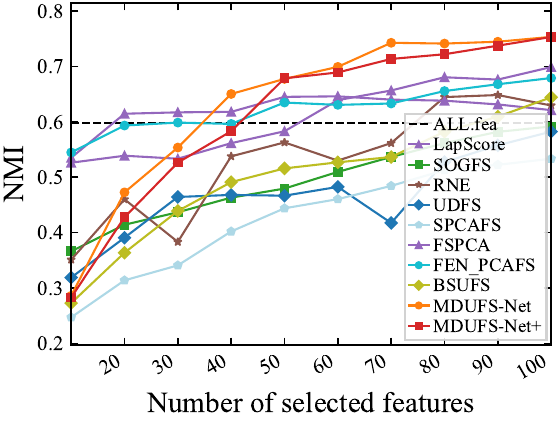}%
\label{fig:ISOLET_NMI}%
}\hspace{1mm}
\subfloat[USPS]{%
\includegraphics[width=4.3cm]{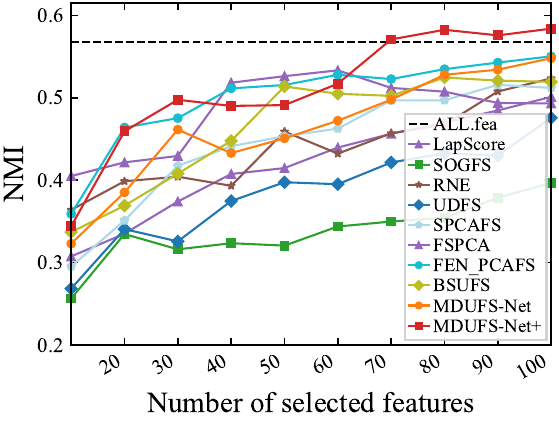}%
\label{fig:USPS_NMI}%
}\hspace{1mm}
\subfloat[PIE]{%
\includegraphics[width=4.3cm]{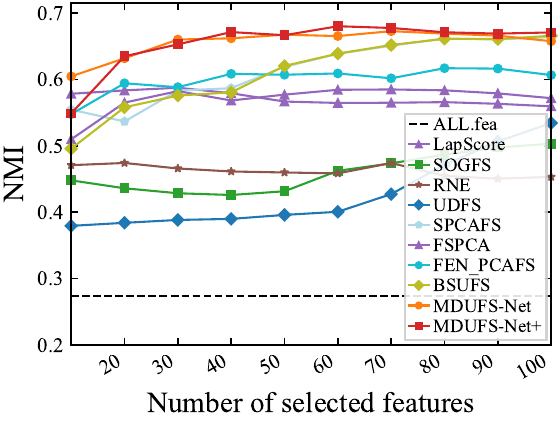}%
\label{fig:PIE_NMI}%
}

\par\vspace{0.5mm}

\subfloat[LUNG]{%
\includegraphics[width=4.3cm]{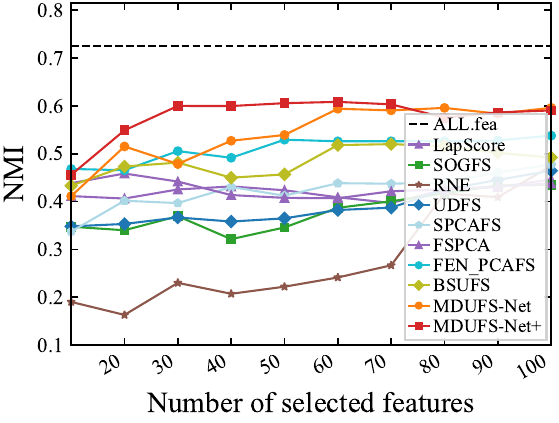}%
\label{fig:LUNG_NMI}%
}\hspace{1mm}
\subfloat[MSTAR]{%
\includegraphics[width=4.3cm]{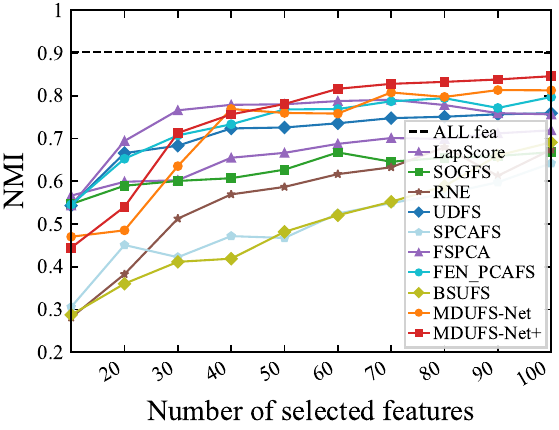}%
\label{fig:MSTAR_NMI}%
}\hspace{1mm}
\subfloat[MNIST]{%
\includegraphics[width=4.3cm]{figure/main_pic/NMI_MSTAR.pdf}%
\label{fig:MNIST_NMI}%
}\hspace{1mm}
\subfloat[HAR]{%
\includegraphics[width=4.3cm]{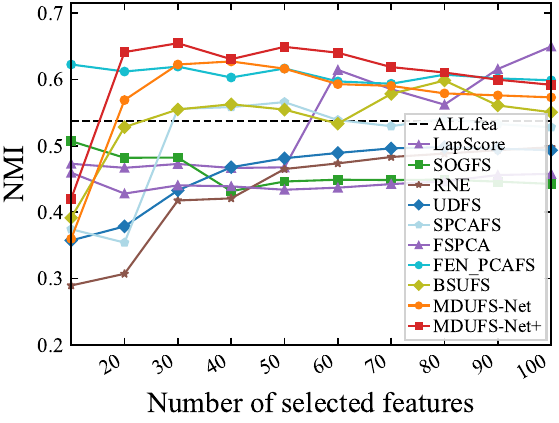}%
\label{fig:HAR_NMI}%
}

\caption{Visual comparisons of the NMI metric under different real-world datasets with different numbers of selected features.}
\label{NMI}

\end{figure*}
\section{Numerical Studies}\label{experiment}

To evaluate the performance of our proposed MDUFS-Net and MDUFS-Net+, extensive experiments are conducted against several representative UFS methods, including LapScore \cite{he2005laplacian}, SOGFS \cite{nie2016unsupervised}, RNE \cite{liu2020robust}, UDFS \cite{yang2011l21}, SPCAFS \cite{li2023sparse}, FSPCA \cite{nie2023learning}, FEN-PCAFS \cite{gao2024principal}, and BSUFS \cite{xiu2025bi}. The source codes of LapScore, SOGFS, RNE, and UDFS are obtained from the AutoUFSTool toolbox\footnote{\url{https://github.com/farhadabedinzadeh/AutoUFSTool} \label{AutoUFSTool}}. For SPCAFS\footnote{\url{https://github.com/quiter2005/algorithm}}, FSPCA\footnote{\url{https://github.com/tianlai09/FSPCA}}, FEN-PCAFS\footnote{\url{https://github.com/gaoyl-group/FEN-PCAFS}}, and BSUFS\footnote{\url{https://github.com/xianchaoxiu/BSUFS}}, the publicly available implementations released by the corresponding authors are adopted.

Subsection \ref{setup} describes the datasets, parameter settings, and evaluation metrics. Subsections \ref{experiments} and \ref{experiments-real} present the experimental results on synthetic and real-world datasets, respectively. Subsection \ref{ablation} reports the ablation experiments. Subsection \ref{discussion} provides further discussions.

\subsection{Experimental Setup}\label{setup}
\subsubsection{Dataset Descriptions}

Two synthetic datasets and eight real-world datasets are used to evaluate the performance of the proposed methods, as summarized in Table \ref{data}.
Specifically, the  synthetic datasets\footnote{\url{https://github.com/milaan9/Clustering-Datasets}} are generated by assigning specific distributions to the first two features, while the remaining seven features are filled with Gaussian noise. The eight real-world datasets cover a variety of application domains, where Isolet\footnote{\url{https://jundongl.github.io/scikit-feature/datasets.html} \label{web-data}} for spoken letter recognition, MSTAR\_SOC\_CNN\footnote{\url{https://github.com/zjj20212035/SPCA-PSD}} (referred to as MSTAR) for deep learning, lung\_discrete\textsuperscript{\ref{web-data}} (referred to as LUNG) for biological information, COIL20\textsuperscript{\ref{web-data}}, USPS\textsuperscript{\ref{web-data}}, PIE\footnote{\url{https://data.nvision2.eecs.yorku.ca/PIE\_dataset/}}, and MNIST\footnote{\url{https://github.com/zalandoresearch/fashion-mnist/}} for image processing,  HAR\footnote{\url{https://archive.ics.uci.edu/dataset/}} for time-series analysis.

\subsubsection{Parameter Settings}\label{parameter setting}

For LapScore, SOGFS, UDFS, and RNE, the value of $k$-neighbors is determined as 5. For SOGFS, SPCAFS, FEN-PCAFS, and BSUFS, the regularization parameters are selected from the candidate set $\{10^{-6}, 10^{-4}, 10^{-2}, 10^{0}, 10^{2}, 10^{4}, 10^{6}\}$.
To ensure a fair comparison, the proposed MDUFS-Net and MDUFS-Net+ use the same network architecture and training strategy across all datasets. The number of unfolding stages is set to 15, and the models are trained for 100 epochs using the Adam optimizer with an initial learning rate of $1 \times 10^{-4}$. The learnable parameters are properly initialized and optimized automatically through gradient-based learning during training. The lower bounds of sparse regularization parameters are set according to the feature dimensionality of each dataset.
\subsubsection{Evaluation Metrics}
Two key metrics are applied to evaluate these compared methods, including clustering accuracy (ACC) and normalized mutual information (NMI) \cite{vinh2009information}. Following a similar setting in \cite{xiu2025bi}, the number of selected features is varied from 10 to 100 in increments of 10 across all datasets. To reduce the variability caused by different initial conditions, the $k$-means clustering algorithm is independently repeated 50 times. The final results are reported in terms of the mean and standard deviation over these 50 runs.

%TABLE2==========================================
\begin{table*}[t]
\caption{Average ACC (mean \% $\pm$ std \%) and number of selected features
(in brackets) of $k$-means clustering,\\where the best results are marked in bold.}
\label{acc}
\centering
\renewcommand{\arraystretch}{1.1}
\footnotesize
\setlength{\tabcolsep}{1.5mm}

%================ 单一大表 =================
\begin{tabular}{|c|c|c|c|c|c|c|c|c|c|}
\hline

{Methods}
& {COIL20}
& {ISOLET}
& {USPS}
& {PIE}
& {LUNG}
& {MSTAR}
& {MNIST}
& {HAR}
& {Average}
\\

\hline\hline

% ==================== ALLfea ====================
\multirow{2}{*}{ALLfea}
& 64.73$\pm$7.63
& 59.87$\pm$2.78
& 59.82$\pm$1.29
& 27.43$\pm$0.96
& 72.57$\pm$7.63
& 90.29$\pm$3.94
& 47.69$\pm$2.78
& 53.79$\pm$0.99
& \multirow{2}{*}{59.52$\pm$3.50}
\\

& (10)
& (10)
& (10)
& (10)
& (10)
& (10)
& (10)
& (10)
&
\\
\hline

% ==================== LapScore ====================
\multirow{2}{*}{LapScore}
& 55.36$\pm$1.58
& 54.07$\pm$1.80
& 52.31$\pm$2.68
& 33.90$\pm$0.97
& 58.40$\pm$0.75
& 68.67$\pm$3.32
& 48.62$\pm$0.15
& 61.59$\pm$0.09
& \multirow{2}{*}{54.12$\pm$1.42}
\\

& (90)
& (100)
& (100)
& (30)
& (100)
& (100)
& (80)
& (100)
&
\\
\hline

% ==================== SOGFS ====================
\multirow{2}{*}{SOGFS}
& 43.32$\pm$2.96
& 43.43$\pm$1.84
& 45.23$\pm$1.31
& 24.73$\pm$0.94
& 54.02$\pm$3.19
& 64.65$\pm$2.66
& 41.26$\pm$1.15
& 46.05$\pm$0.38
& \multirow{2}{*}{45.34$\pm$1.80}
\\

& (100)
& (30)
& (100)
& (100)
& (100)
& (60)
& (80)
& (100)
&
\\
\hline

% ==================== RNE ====================
\multirow{2}{*}{RNE}
& 56.64$\pm$3.40
& 50.29$\pm$1.77
& 57.51$\pm$2.01
& 22.39$\pm$0.61
& 58.90$\pm$5.79
& 69.11$\pm$1.02
& 39.58$\pm$0.19
& 42.16$\pm$0.03
& \multirow{2}{*}{49.57$\pm$1.85}
\\

& (80)
& (100)
& (100)
& (60)
& (100)
& (80)
& (100)
& (60)
&
\\
\hline

% ==================== UDFS ====================
\multirow{2}{*}{UDFS}
& 48.38$\pm$2.41
& 40.84$\pm$1.83
& 50.63$\pm$1.93
& 27.49$\pm$0.94
& 58.19$\pm$5.29
& 75.24$\pm$3.37
& 34.57$\pm$1.37
& 42.03$\pm$0.03
& \multirow{2}{*}{47.17$\pm$2.15}
\\

& (100)
& (100)
& (100)
& (100)
& (100)
& (90)
& (80)
& (100)
&
\\
\hline

% ==================== SPCAFS ====================
\multirow{2}{*}{SPCAFS}
& 45.72$\pm$2.07
& 40.55$\pm$1.76
& 57.95$\pm$1.57
& 45.41$\pm$1.74
& 56.32$\pm$4.29
& 67.96$\pm$2.81
& 31.71$\pm$0.91
& 58.67$\pm$0.01
& \multirow{2}{*}{50.54$\pm$1.90}
\\

& (80)
& (100)
& (80)
& (80)
& (90)
& (100)
& (90)
& (90)
&
\\
\hline

% ==================== FSPCA ====================
\multirow{2}{*}{FSPCA}
& 56.57$\pm$1.89
& 47.99$\pm$1.31
& 56.05$\pm$1.96
& 32.45$\pm$1.18
& 55.74$\pm$1.90
& 80.09$\pm$5.55
& 49.49$\pm$1.43
& 44.54$\pm$0.05
& \multirow{2}{*}{52.87$\pm$1.91}
\\

& (80)
& (50)
& (60)
& (80)
& (90)
& (70)
& (70)
& (40)
&
\\
\hline

% ==================== FEN-PCAFS ====================
\multirow{2}{*}{FEN-PCAFS}
& 58.20$\pm$2.24
& 52.28$\pm$2.11
& 58.28$\pm$2.31
& 38.94$\pm$1.64
& 61.77$\pm$1.53
& 81.07$\pm$3.84
& 47.66$\pm$0.02
& 60.91$\pm$0.04
& \multirow{2}{*}{57.39$\pm$1.72}
\\

& (100)
& (100)
& (100)
& (80)
& (80)
& (80)
& (60)
& (20)
&
\\
\hline

% ==================== BSUFS ====================
\multirow{2}{*}{BSUFS}
& 57.26$\pm$2.37
& 53.59$\pm$2.12
& 59.33$\pm$0.96
& 45.25$\pm$1.74
& 60.13$\pm$2.42
& 71.45$\pm$1.81
& 33.80$\pm$0.89
& 63.27$\pm$0.05
& \multirow{2}{*}{55.51$\pm$1.55}
\\

& (100)
& (100)
& (80)
& (80)
& (70)
& (100)
& (100)
& (90)
&
\\
\hline

% ==================== MDUFS-Net ====================

\rowcolor{gray!10}&\textbf{65.60$\pm$2.83}
& 62.84$\pm$2.27
& 62.02$\pm$1.69
& 45.61$\pm$1.73
& 66.12$\pm$1.86
& 88.00$\pm$1.83
& 56.10$\pm$0.05
& 67.53$\pm$0.25
&
\\

\rowcolor{gray!10}\multirow{-2}{*}{MDUFS-Net}
& \textbf{(100)}
& (100)
& (100)
& (70)
& (100)
& (90)
& (100)
& (40)
& \multirow{-2}{*}{64.23$\pm$1.56}
\\ \hline

% ================= MDUFS-Net+==================

\rowcolor{gray!10}&65.23$\pm$1.89
& \textbf{64.16$\pm$2.04}
& \textbf{64.79$\pm$1.65}
& \textbf{46.82$\pm$1.65}
& \textbf{69.46$\pm$1.35}
& \textbf{91.70$\pm$0.15}
& \textbf{56.75$\pm$0.03}
& \textbf{70.19$\pm$0.13}
&
\\

\rowcolor{gray!10}\multirow{-2}{*}{MDUFS-Net+}
& (90)
& \textbf{(100)}
& \textbf{(80)}
& \textbf{(60)}
& \textbf{(60)}
& \textbf{(100)}
& \textbf{(100)}
& \textbf{(20)}
& \multirow{-2}{*}{\textbf{66.14$\pm$1.11}}
\\
\hline

\end{tabular}

\end{table*}
%==================================================
%======================TABLE3=====================
\begin{table*}[!t]

\caption{Average NMI (mean \% $\pm$ std \%) and number of selected features
(in brackets) of $k$-means clustering,\\ where best results are marked in bold.}
\label{nmi}
\centering
\renewcommand{\arraystretch}{1.1}
\footnotesize
\setlength{\tabcolsep}{1.5mm}
\begin{tabular}{|c|c|c|c|c|c|c|c|c|c|}
\hline
{Methods}
& {COIL20}
& {ISOLET}
& {USPS}
& {PIE}
& {LUNG}
& {MSTAR}
& {MNIST}
& {HAR}
& {Average}
\\
\hline\hline

% ==================== ALLfea ====================
\multirow{2}{*}{ALLfea}
& 77.43$\pm$1.10
& 74.57$\pm$1.40
& 56.73$\pm$1.10
& 52.42$\pm$0.92
& 60.04$\pm$3.59
& 87.39$\pm$1.10
& 50.65$\pm$0.26
& 55.63$\pm$1.67
& \multirow{2}{*}{64.36$\pm$1.39}
\\

& (10)
& (10)
& (10)
& (10)
& (10)
& \textbf{(10)}
& (10)
& (10)
&
\\
\hline

% ================== LapScore ==================
\multirow{2}{*}{LapScore}
& 69.47$\pm$0.85
& 69.92$\pm$0.83
& 50.10$\pm$1.54
& 58.74$\pm$0.63
& 44.62$\pm$0.99
& 71.60$\pm$0.46
& 50.67$\pm$0.22
& 64.95$\pm$0.09
& \multirow{2}{*}{60.01$\pm$0.70}
\\

& (80)
& (100)
& (100)
& (30)
& (100)
& (100)
& (80)
& (100)
&
\\
\hline

% ==================== SOGFS ====================
\multirow{2}{*}{SOGFS}
& 60.42$\pm$1.40
& 59.18$\pm$1.04
& 39.67$\pm$1.41
& 50.31$\pm$0.88
& 43.50$\pm$3.32
& 66.78$\pm$0.65
& 40.49$\pm$1.15
& 48.22$\pm$0.01
& \multirow{2}{*}{51.07$\pm$1.23}
\\

& (100)
& (30)
& (100)
& (100)
& (100)
& (100)
& (100)
& (30)
&
\\
\hline

% ==================== RNE ====================
\multirow{2}{*}{RNE}
& 70.57$\pm$1.55
& 64.87$\pm$0.89
& 52.35$\pm$1.15
& 47.41$\pm$0.62
& 47.46$\pm$4.07
& 67.38$\pm$1.91
& 30.27$\pm$0.21
& 49.78$\pm$0.01
& \multirow{2}{*}{53.76$\pm$1.30}
\\

& (80)
& (90)
& (100)
& (20)
& (100)
& (80)
& (100)
& (100)
&
\\
\hline

% ==================== UDFS ====================
\multirow{2}{*}{UDFS}
& 69.57$\pm$1.07
& 58.26$\pm$1.35
& 47.56$\pm$1.28
& 53.42$\pm$0.80
& 46.45$\pm$3.57
& 75.89$\pm$0.67
& 36.71$\pm$0.33
& 49.33$\pm$0.03
& \multirow{2}{*}{54.65$\pm$1.14}
\\

& (100)
& (100)
& (100)
& (100)
& (100)
& (90)
& (100)
& (100)
&
\\
\hline

% ================== SPCAFS ====================
\multirow{2}{*}{SPCAFS}
& 61.78$\pm$1.12
& 53.36$\pm$1.03
& 51.56$\pm$0.53
& 66.36$\pm$0.91
& 47.73$\pm$2.86
& 64.20$\pm$2.27
& 30.72$\pm$1.28
& 56.57$\pm$0.15
& \multirow{2}{*}{54.04$\pm$1.27}
\\

& (100)
& (100)
& (90)
& (70)
& (100)
& (100)
& (90)
& (50)
&
\\
\hline

% ==================== FSPCA ====================
\multirow{2}{*}{FSPCA}
& 73.24$\pm$0.86
& 64.54$\pm$0.66
& 53.33$\pm$0.89
& 58.47$\pm$0.67
& 43.79$\pm$1.71
& 79.06$\pm$2.03
& 50.38$\pm$0.36
& 45.77$\pm$0.22
& \multirow{2}{*}{58.57$\pm$0.93}
\\

& (80)
& (40)
& (60)
& (80)
& (100)
& (100)
& (70)
& (100)
&
\\
\hline

% ================== FEN-PCAFS ===================
\multirow{2}{*}{FEN-PCAFS}
& 73.16$\pm$1.08
& 67.93$\pm$1.04
& 55.03$\pm$1.27
& 61.68$\pm$1.16
& 53.78$\pm$0.79
& 79.70$\pm$1.47
& 47.66$\pm$0.02
& 61.92$\pm$0.01
& \multirow{2}{*}{62.61$\pm$0.86}
\\

& (100)
& (100)
& (100)
& (80)
& (100)
& (80)
& (60)
& (30)
&
\\
\hline

% ==================== BSUFS ====================
\multirow{2}{*}{BSUFS}
& 71.18$\pm$1.39
& 64.46$\pm$1.35
& 52.46$\pm$0.64
& 66.55$\pm$0.94
& 52.00$\pm$1.48
& 69.09$\pm$1.06
& 33.80$\pm$0.89
& 59.83$\pm$0.12
& \multirow{2}{*}{58.67$\pm$0.98}
\\

& (100)
& (100)
& (80)
& (100)
& (70)
& (100)
& (100)
& (80)
&
\\
\hline

% ================= MDUFS-Net===================
 
\rowcolor{gray!10}&78.63$\pm$1.20
& \textbf{75.38$\pm$0.87}
& 54.80$\pm$1.60
& 67.26$\pm$1.10
& 59.60$\pm$1.46
& 81.32$\pm$1.18
& 50.27$\pm$0.02
& 62.70$\pm$0.09
&
\\
 
\rowcolor{gray!10}\multirow{-2}{*}{MDUFS-Net}
& (100)
& \textbf{(100)}
& (100)
& (70)
& (100)
& (90)
& (100)
& (40)
& \multirow{-2}{*}{66.25$\pm$0.94}
\\
\hline

% ================ MDUFS-Net+===================
 
\rowcolor{gray!10}&\textbf{78.96$\pm$1.02}
& 75.36$\pm$1.16
& \textbf{58.39$\pm$0.54}
& \textbf{68.00$\pm$0.79}
& \textbf{60.84$\pm$1.42}
& \textbf{84.57$\pm$0.21}
& \textbf{50.86$\pm$0.02}
& \textbf{65.44$\pm$0.65}
&
\\
 
\rowcolor{gray!10}\multirow{-2}{*}{MDUFS-Net+}
& \textbf{(90)}
& (100)
& \textbf{(100)}
& \textbf{(60)}
& \textbf{(60)}
& (100)
& \textbf{(90)}
& \textbf{(30)}
& \multirow{-2}{*}{\textbf{67.80$\pm$0.73}}
\\
\hline

\end{tabular}

\end{table*}
%================================================

\subsection{Synthetic Results}\label{experiments}

In this experiment, all compared methods are evaluated on two synthetic datasets. For each method, features are ranked according to their selection scores, and the top two features are selected. Afterwards, the selected features and corresponding sample distributions are visualized using scatter plots.

Fig. \ref{dartboard1} shows that for the Dartboard1 dataset, LapScore, UDFS, FEN-PCAFS, MDUFS-Net, and MDUFS-Net+ are capable of identifying the appropriate features, compared with other methods. Furthermore, Fig. \ref{Diamond9} demonstrates that for the Diamond9 dataset, only MDUFS-Net and MDUFS-Net+ successfully identify the two most discriminative features, while the compared methods fail to accurately identify them. Overall, our MDUFS-Net and MDUFS-Net+ consistently yield reliable feature selection results, which validate the effectiveness of the proposed methods on synthetic datasets.

% ================= TABLE 4 ====================
\begin{table*}[!t]

\caption{Ablation studies of sparsity regularization terms, where
the best results are marked in bold.}
\label{ablation_sparity}

\centering
\renewcommand{\arraystretch}{1.1}
\footnotesize
\setlength{\tabcolsep}{3.5mm}
\begin{tabular}{|c|c|c|c|c|c|c|c|c|c|c|}
\hline

{$\ell_{2,0}$-norm}
& {$\ell_{0}$-norm}
& {Metrics}
& {COIL20}
& {ISOLET}
& {USPS}
& {PIE}
& {LUNG}
& {MSTAR}
& {MNIST}
& {HAR}
\\

\hline\hline

% ==================== Case I ====================
\multirow{2}{*}{$\times$}
& \multirow{2}{*}{$\times$}
& ACC
& 44.42
& 40.20
& 61.62
& 45.46
& 63.09
& 81.46
& 42.22
& 45.16
\\

&
& NMI
& 63.16
& 52.00
& 53.52
& 66.37
& 58.51
& 79.76
& 41.99
& 54.13
\\
\hline

% ==================== Case II ====================
\multirow{2}{*}{$\times$}
& \multirow{2}{*}{$\checkmark$}
& ACC
& 45.34
& 40.31
& 59.21
& 44.52
& 60.43
& 81.48
& 41.83
& 50.93
\\

&
& NMI
& 62.88
& 51.79
& 52.91
& 65.87
& 57.64
& 79.66
& 42.06
& 54.14
\\
\hline

% ==================== Case III ====================
 \multirow{2}{*}{$\checkmark$}
& \multirow{2}{*}{$\times$}
& ACC
& 63.79
& 62.84
& 66.58
& 44.76
& 65.18
& 87.05
& 56.10
& 50.60
\\

&
& NMI
& 77.27
& \textbf{75.38}
& 57.29
& 66.52
& 59.40
& 81.26
& 49.66
& 62.24
\\
\hline

% ==================== Case IV ====================
 \multirow{2}{*}{$\checkmark$}
& \multirow{2}{*}{$\checkmark$}
& ACC
& \textbf{65.23}
& \textbf{64.16}
& \textbf{68.35}
& \textbf{46.82}
& \textbf{69.46}
& \textbf{91.70}
& \textbf{56.75}
& \textbf{70.19}
\\

&
& NMI
& \textbf{78.96}
& 75.36
& \textbf{59.10}
& \textbf{68.00}
& \textbf{60.84}
& \textbf{84.57}
& \textbf{50.25}
& \textbf{65.44}
\\
\hline

\end{tabular}

\end{table*}
% ================= TABLE 4 END==================

% ================ TABLE 6 ==================
\begin{table*}[!t]

\caption{Ablation studies of manifold optimization, where
the best results are marked in bold.}
\label{ablation_manifold}

\centering
\renewcommand{\arraystretch}{1.1}
\footnotesize
\setlength{\tabcolsep}{3.4mm}

\begin{tabular}{|c|c|c|c|c|c|c|c|c|c|c|}
\hline

{Projection}
& {Retraction}
& {Metrics}
& {COIL20}
& {ISOLET}
& {USPS}
& {PIE}
& {LUNG}
& {MSTAR}
& {MNIST}
& {HAR}
\\

\hline\hline

% ================== Case I ====================
\multirow{2}{*}{$\times$}
& \multirow{2}{*}{$\times$}
& ACC
& 56.23
& 21.57
& 44.30
& 20.49
& 49.67
& 52.15
& 42.39
& 37.77
\\

&
& NMI
& 66.51
& 36.13
& 39.23
& 41.77
& 38.52
& 55.68
& 41.55
& 47.47
\\
\hline

% ================= Case II ====================
\multirow{2}{*}{$\times$}
& \multirow{2}{*}{$\checkmark$}
& ACC
& 54.02
& 49.91
& 49.85
& 32.77
& 50.81
& 72.55
& 43.54
& 53.53
\\

&
& NMI
& 68.86
& 61.94
& 46.18
& 59.33
& 41.23
& 73.08
& 45.22
& 52.64
\\
\hline

% ================= Case III ====================
 \multirow{2}{*}{$\checkmark$}
& \multirow{2}{*}{$\times$}
& ACC
& 59.48
& 49.71
& 60.04
& 32.31
& 58.39
& 78.83
& 49.80
& 49.99
\\

&
& NMI
& 72.21
& 62.53
& 54.23
& 56.35
& 49.49
& 74.52
& 47.02
& 49.03
\\
\hline

% ================= Case IV ====================
 \multirow{2}{*}{$\checkmark$}
& \multirow{2}{*}{$\checkmark$}
& ACC
& \textbf{65.23}
& \textbf{64.16}
& \textbf{68.35}
& \textbf{46.82}
& \textbf{69.46}
& \textbf{91.70}
& \textbf{56.75}
& \textbf{70.19}
\\

&
& NMI
& \textbf{78.96}
& \textbf{75.36}
& \textbf{59.10}
& \textbf{68.00}
& \textbf{60.84}
& \textbf{84.57}
& \textbf{50.25}
& \textbf{65.44}
\\
\hline

\end{tabular}

\end{table*}

\subsection{Real-world Results}\label{experiments-real}

In this section, numerical experiments are conducted on eight real-world datasets to evaluate the effectiveness of the proposed methods. Besides, ALLfea is introduced to serve as a baseline for comparison.

Regarding the ACC metric, the performance variations of different methods with respect to the number of selected features are illustrated in Fig. \ref{ACC}. It can be observed that our MDUFS-Net and MDUFS-Net+ generally maintain higher ACC curves than other comparison methods under different feature selection settings. The results in Table \ref{acc} further verify this observation, where MDUFS-Net+ achieves the highest average ACC in most cases, followed by MDUFS-Net. The advantage mainly stems from the deep unfolding framework, which transforms the iterative ADMM optimization into a trainable network and enables adaptive parameters at different stages. Meanwhile, the orthogonality constraint helps preserve the structural information of the feature space, leading to more discriminative feature representations.  Furthermore, compared with MDUFS-Net, MDUFS-Net+ increases the average ACC by 1.91\% and decreases the average standard deviation from 1.56\% to 1.11\%, demonstrating improved clustering performance and stability.

For the NMI metric, similar results can be observed, where the proposed MDUFS-Net and MDUFS-Net+ generally outperform other competitors. As shown in Fig. \ref{NMI}, the curves of MDUFS-Net and MDUFS-Net+ are overall higher than those of the other methods, indicating that the proposed methods achieve better performance in terms of clustering consistency. In Table \ref{nmi}, compared with the best-performing comparison method, MDUFS-Net and MDUFS-Net+ improve the average NMI by 3.64\% and 5.19\%, respectively. It is worth noting that the relatively lower NMI values of MDUFS-Net on the MNIST and HAR datasets may be attributed to the strong feature correlations and overlapping class distributions in these datasets. In contrast, our MDUFS-Net+ benefits from the introduced double sparsity structure, which helps preserve more discriminative information.

\subsection{Ablation Experiments} \label{ablation}

\subsubsection{Effect of Sparsity Regularization Terms} 
In this part, ablation studies are conducted on MDUFS-Net+ in problem \eqref{23} with four different combinations of sparsity regularization terms considered.
As shown in Table \ref{ablation_sparity}, MDUFS-Net+ incorporating both the $\ell_{2,0}$-norm and $\ell_0$-norm regularization terms achieves the best performance across all datasets in terms of ACC. Specifically, on the COIL20 and ISOLET datasets, the ACC values improve from 44.42\% to 65.23\% and from 40.20\% to 64.16\%, respectively. Compared with the settings that incorporate only one type of sparsity regularization term, the complete model demonstrates superior clustering performance in the vast majority of cases, indicating that row sparsity and element-wise sparsity are complementary.

\subsubsection{Effect of Manifold Optimization} 
To investigate the contribution of the proposed manifold mechanism, ablation studies are compared by removing the projection operation and QR-based retraction. As reported in Table \ref{ablation_manifold}, MDUFS-Net+ incorporating the complete manifold mechanism achieves the best performance on all eight datasets. In contrast, removing either the projection operation or the QR-based retraction leads to performance degradation, indicating that both components play important roles in maintaining effective feature selection. 

% ================= TABLE 5 ====================
\begin{table}[!t]

\caption{Ablation studies of the network, where the best results are marked in bold.}
\label{ablation_unfolding}

\centering
\renewcommand{\arraystretch}{1.1}
\footnotesize
\setlength{\tabcolsep}{1.5mm}

\begin{tabular}{|c|c|c|c|c|c|c|}
\hline

{Network}
& {Metrics}
& {COIL20}
& {ISOLET}
& {PIE}
& {LUNG}
& {MNIST}

\\

\hline\hline

% ==================== Without Network ====================
\multirow{2}{*}{$\times$}
& ACC
& 60.81
& 55.70
& 40.02
& 58.83
& 49.91

\\

& NMI
& 74.77
& 70.27
& 64.81
& 49.78
& 48.96

\\
\hline

% ==================== With Network ====================
\multirow{2}{*}{$\checkmark$}
& ACC
& \textbf{65.23}
& \textbf{64.16}
& \textbf{46.82}
& \textbf{69.46}
& \textbf{56.75}

\\

& NMI
& \textbf{78.96}
& \textbf{75.36}
& \textbf{68.00}
& \textbf{60.84}
& \textbf{50.86}

\\
\hline

\end{tabular}

\end{table}
% ================ TABLE 5 END ==================

% ================= TABLE 7 ====================
\begin{table}[t]

\caption{Ablation studies of parameter learning, where the best results are marked in bold.}
\label{ablation_parameter}

\centering
\renewcommand{\arraystretch}{1.1}
\footnotesize
\setlength{\tabcolsep}{1.5mm}

\begin{tabular}{|c|c|c|c|c|c|c|}
\hline

{Strategy}
& {Metrics}
& {COIL20}
& {ISOLET}
& {PIE}
& {LUNG}
& {MNIST}

\\

\hline\hline

% ==================== shared ====================
\multirow{2}{*}{$\times$}
& ACC
& 61.26
& 53.68
& 43.14
& 61.27
& 51.77

\\

& NMI
& 74.56
& 68.35
& 66.08
& 55.14
& 47.44

\\
\hline

% ==================== united ====================
\multirow{2}{*}{$\checkmark$}
& ACC
& \textbf{65.23}
& \textbf{64.16}
& \textbf{46.82}
& \textbf{69.46}
& \textbf{56.75}

\\

& NMI
& \textbf{78.96}
& \textbf{75.36}
& \textbf{68.00}
& \textbf{60.84}
& \textbf{50.86}

\\
\hline

\end{tabular}

\end{table}
% ================ TABLE 5 END ==================

\begin{figure*}[t]
\centering

\subfloat[COIL20]{%
\includegraphics[width=4.3cm]{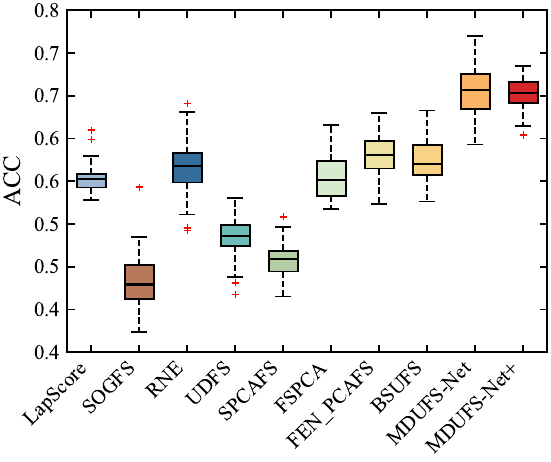}%
\label{COIL20_staility}%
}\hspace{1mm}
\subfloat[ISOLET]{%
\includegraphics[width=4.3cm]{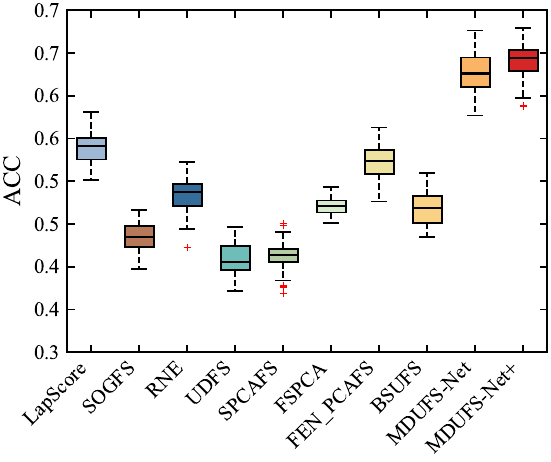}%
\label{ISOLET_staility}%
}\hspace{1mm}
\subfloat[USPS]{%
\includegraphics[width=4.3cm]{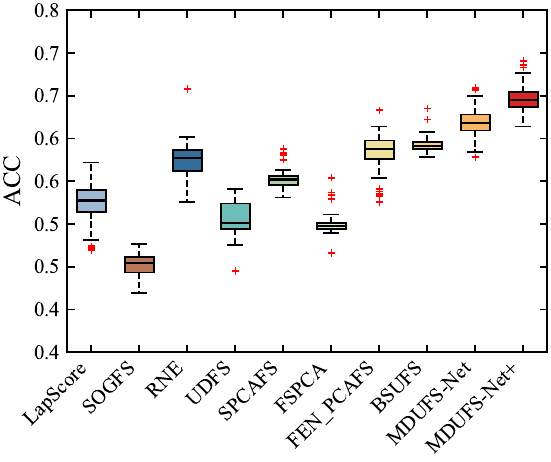}%
\label{USPS_staility}%
}\hspace{1mm}
\subfloat[PIE]{%
\includegraphics[width=4.3cm]{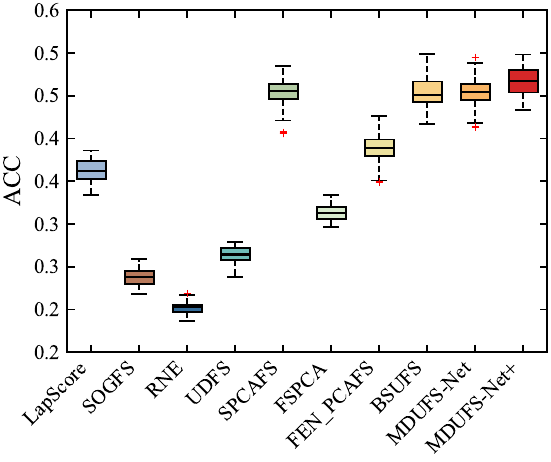}%
\label{PIE_staility}%
}

\par\vspace{0.5mm}

\subfloat[LUNG]{%
\includegraphics[width=4.3cm]{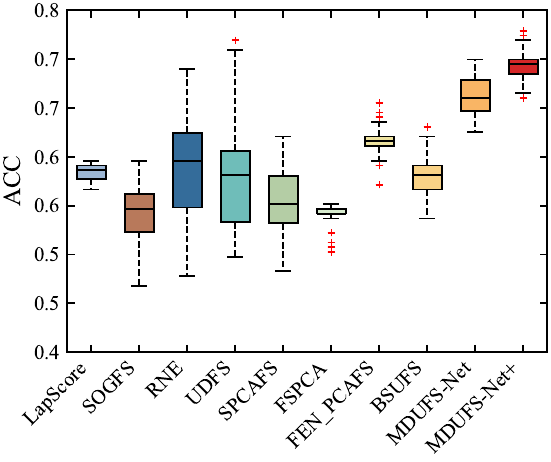}%
\label{LUNG_staility}%
}\hspace{1mm}
\subfloat[MSTAR]{%
\includegraphics[width=4.3cm]{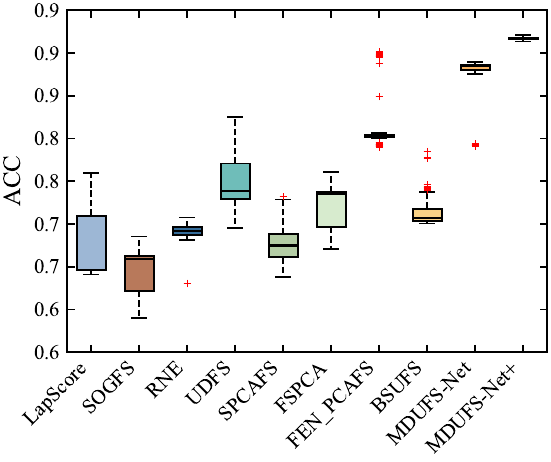}%
\label{MSTAR_staility}%
}\hspace{1mm}
\subfloat[MNIST]{%
\includegraphics[width=4.3cm]{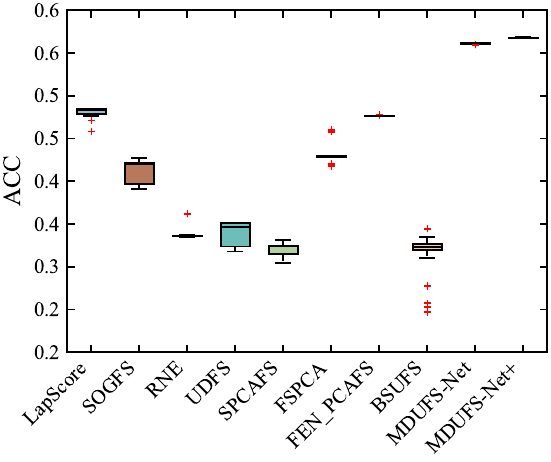}%
\label{MNIST_staility}%
}\hspace{1mm}
\subfloat[HAR]{%
\includegraphics[width=4.3cm]{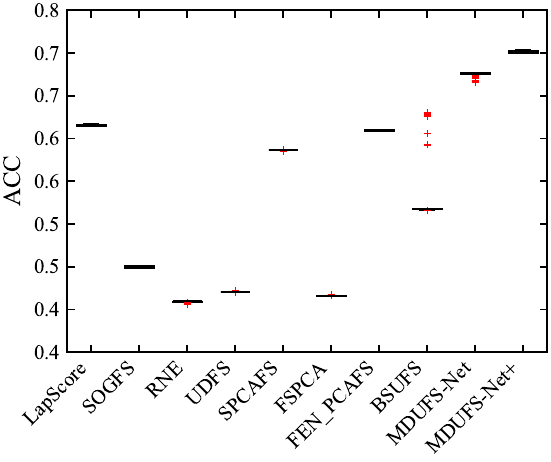}%
\label{HAR_staility}%
}

\caption{Model stability comparisons of all compared methods on eight real-world datasets.}
\label{box}

\end{figure*}

\begin{figure*}[t]
\centering

\subfloat[COIL20]{%
\includegraphics[width=4.3cm]{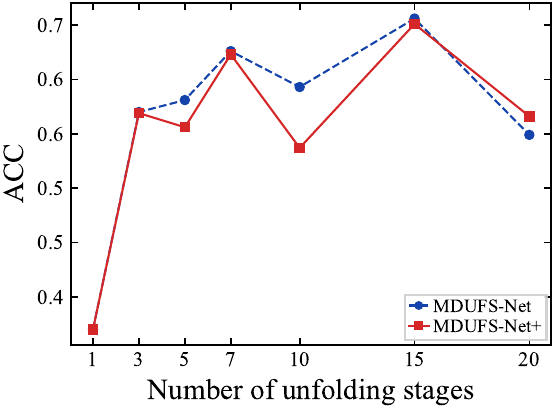}%
\label{COIL20_stage}%
}\hspace{1mm}
\subfloat[ISOLET]{%
\includegraphics[width=4.3cm]{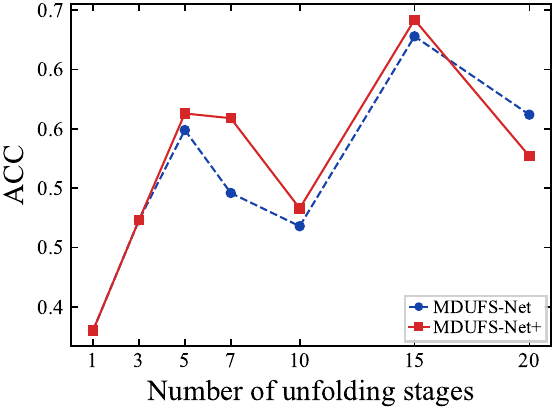}%
\label{ISOLET_stage}%
}\hspace{1mm}
\subfloat[USPS]{%
\includegraphics[width=4.3cm]{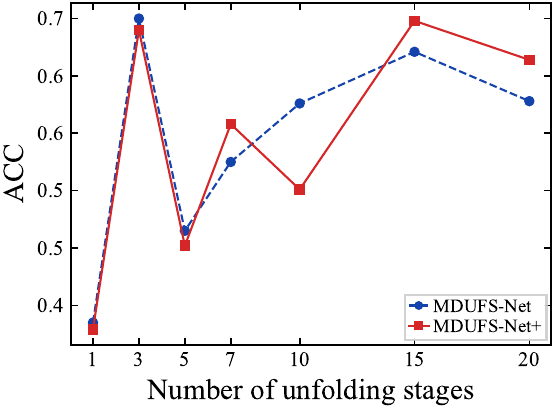}%
\label{USPS_stage}%
}\hspace{1mm}
\subfloat[PIE]{%
\includegraphics[width=4.3cm]{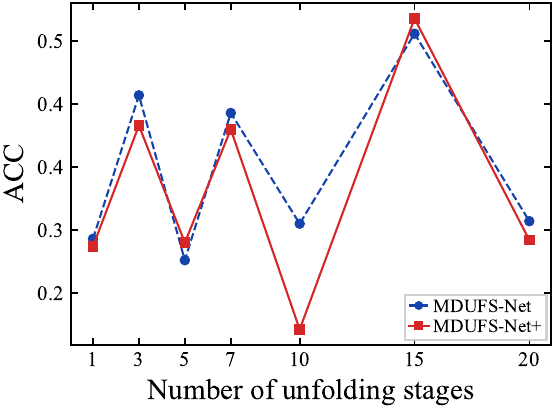}%
\label{PIE_stage}%
}

\par\vspace{0.5mm}

\subfloat[LUNG]{%
\includegraphics[width=4.3cm]{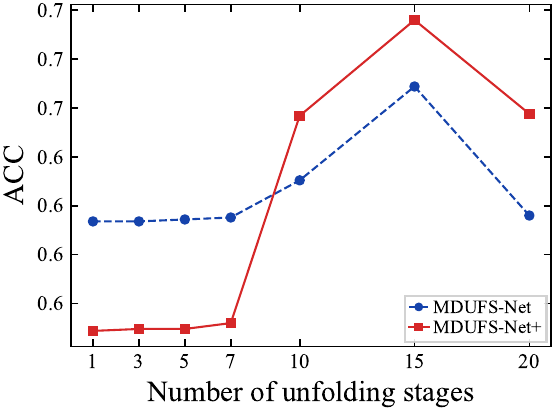}%
\label{LUNG_stage}%
}\hspace{1mm}
\subfloat[MSTAR]{%
\includegraphics[width=4.3cm]{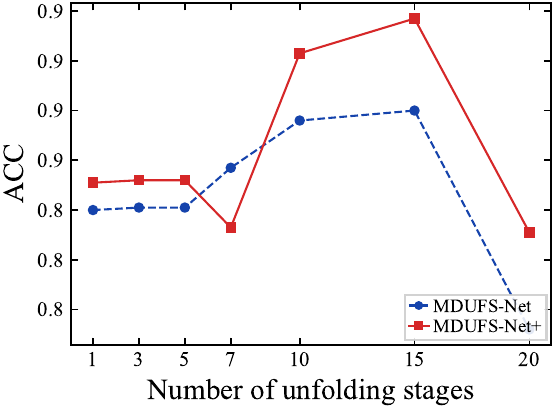}%
\label{MSTAR_stage}%
}\hspace{1mm}
\subfloat[MNIST]{%
\includegraphics[width=4.3cm]{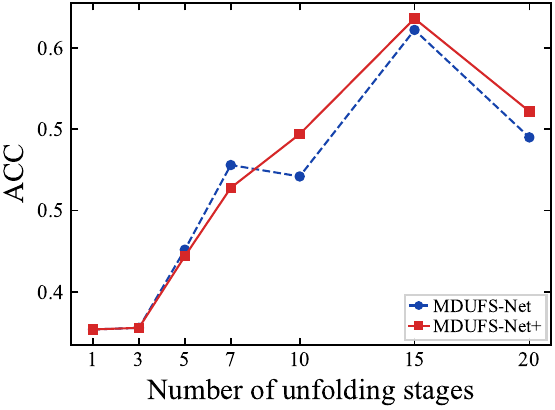}%
\label{MNIST_stage}%
}\hspace{1mm}
\subfloat[HAR]{%
\includegraphics[width=4.3cm]{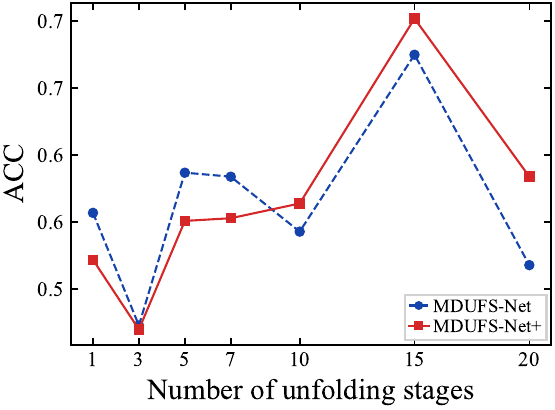}%
\label{HAR_stage}%
}

\caption{Unfolding sensitivity of stages $K$ on eight real-world datasets.}
\label{stage}

\end{figure*}

\begin{figure*}[t]
\centering

\subfloat[COIL20]{%
\includegraphics[width=4.3cm]{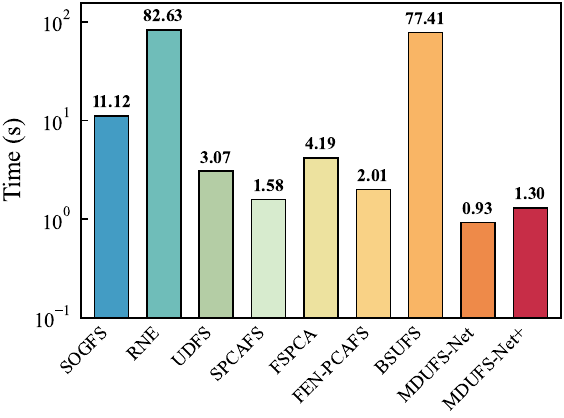}%
\label{time_COIL20}%
}\hspace{1mm}
\subfloat[ISOLET]{%
\includegraphics[width=4.3cm]{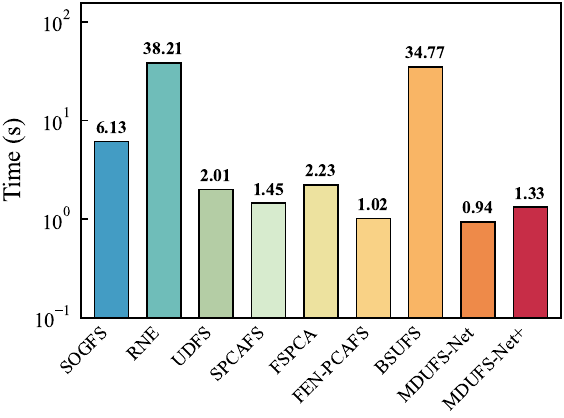}%
\label{time_ISOLET}%
}\hspace{1mm}
\subfloat[USPS]{%
\includegraphics[width=4.3cm]{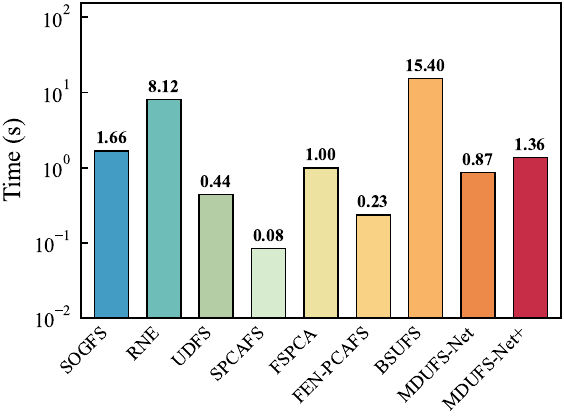}%
\label{time_USPS}%
}\hspace{1mm}
\subfloat[PIE]{%
\includegraphics[width=4.3cm]{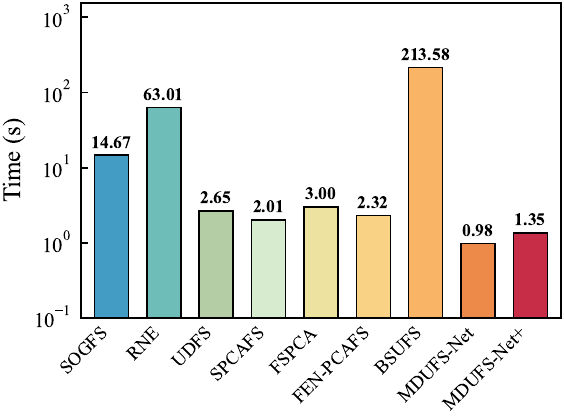}%
\label{time_PIE}%
}

\par\vspace{0.5mm}

\subfloat[LUNG]{%
\includegraphics[width=4.3cm]{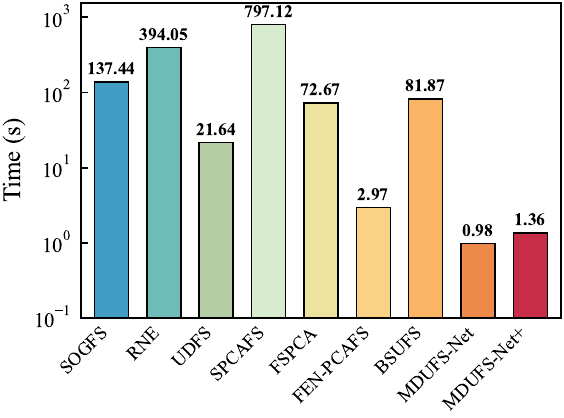}%
\label{time_LUNG}%
}\hspace{1mm}
\subfloat[MSTAR]{%
\includegraphics[width=4.3cm]{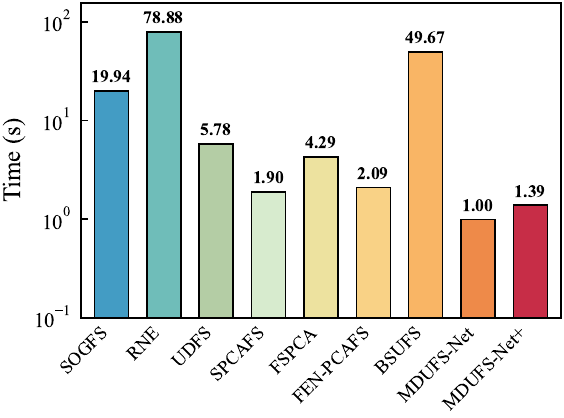}%
\label{time_MSTAR}%
}\hspace{1mm}
\subfloat[MNIST]{%
\includegraphics[width=4.3cm]{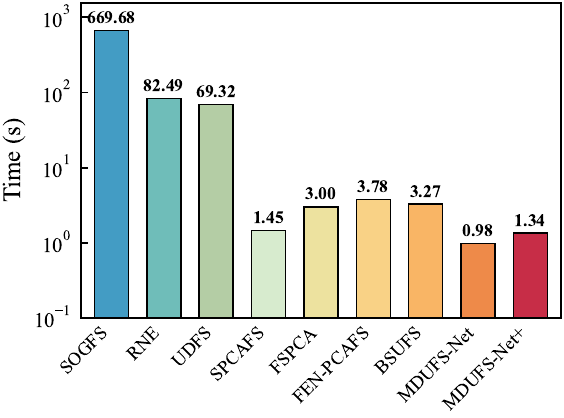}%
\label{time_MNIST}%
}\hspace{1mm}
\subfloat[HAR]{%
\includegraphics[width=4.3cm]{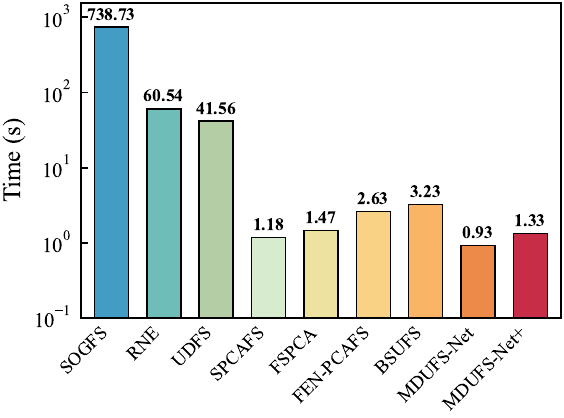}%
\label{time_HAR}%
}

\caption{Runtime comparisons among different methods on eight real-world datasets.}
\label{time}

\end{figure*}

\subsubsection{Effect of Deep Unfolding} 
Table \ref{ablation_unfolding} reports the ablation results of the deep unfolding network in five selected datasets. Here, “$\times$” indicates the conventional ADMM optimization with fixed parameters, while “$\checkmark$” denotes our MDUFS-Net+. The results demonstrate that the integration of the deep unfolding strategy leads to consistent improvements in both ACC and NMI across all datasets. Taking the COIL20 dataset as an example, the ACC is improved from 60.81\% to 65.23\%. Similar gains are observed on the remaining datasets, further validating the effectiveness of the proposed deep unfolding mechanism for feature selection and clustering tasks.

\subsubsection{Effect of Parameter Learning} 
To evaluate parameter learning strategies, ablation studies compare shared and stage-wise independent parameters, where $\times$ and $\checkmark$ denote shared and independent learning, respectively. As shown in Table \ref{ablation_parameter}, stage-wise independent learning consistently outperforms parameter sharing. This improvement benefits from the flexible parameter adaptation at different unfolding stages, validating the effectiveness of stage-wise parameter learning in the proposed deep unfolding framework.

\subsection{Discussions} \label{discussion}
This section presents a further analysis of our MDUFS-Net and MDUFS-Net+ from three aspects, including model stability, unfolding sensitivity, and computational complexity.

\subsubsection{Model Stability}
In this experiment, box plots of 50 clustering results  are presented in Fig. \ref{box}. Our MDUFS-Net and MDUFS-Net+ achieve better ACC performance with more compact result distributions than compared methods. Furthermore, MDUFS-Net+ exhibits improved stability, especially on datasets such as ISOLET and PIE. This improvement is attributed to the $\ell_{0}$-norm, which promotes sparse feature selection and reduces the influence of redundant features.

\subsubsection{Unfolding Sensitivity}
Different numbers of unfolding stages are compared, and the results are shown in Fig. \ref{stage}. As the number of unfolding stages increases, the performance generally tends to improve and gradually stabilizes when sufficient stages are employed. On most datasets, 15 unfolding stages achieve the best or comparable performance, while some datasets can also obtain competitive results with fewer stages, indicating that fewer unfolding stages are sufficient to obtain effective feature selection results for these datasets. Considering the performance and computational cost, 15 unfolding stages are selected as the default setting.

\subsubsection{Runtime}

%Since LapScore has a relatively simple structure and requires much less running time than other methods, it is excluded here. 
Note that the compared methods are implemented in MATLAB and executed on CPU, whereas our methods are implemented in Python with GPU acceleration. As shown in Fig. \ref{time}, the computational costs of different methods vary significantly across datasets. Some optimization-based methods involve iterative optimization procedures, leading to relatively higher runtime. For example, BSUFS and SOGFS exhibit higher computational costs on the PIE and MNIST datasets, respectively, which may be related to factors such as iterative optimization procedures, parameter settings, and implementation environments. In contrast, our proposed MDUFS-Net and MDUFS-Net+ maintain acceptable computational costs across all datasets.

\section{Conclusion}\label{conclusion}

This paper presents two novel UFS methods, MDUFS-Net and MDUFS-Net+, by introducing the $\ell_{2,0}$-norm and $\ell_{0}$-norm into PCA. Specifically, the $\ell_{2,0}$-norm enables feature selection, while the $\ell_{0}$-norm suppresses the influence of redundant feature elements. Moreover, the classical ADMM framework is integrated with deep unfolding to convert the iterative optimization process into a trainable network, thereby reducing the manual parameter tuning required by conventional model-driven methods. Furthermore, the incorporation of the manifold optimization mechanism enhances the stability of the network. Therefore, the proposed methods preserve the advantages of both sparse modeling and deep learning. Experimental results demonstrate that our MDUFS-Net and MDUFS-Net+ achieve effective feature selection performance  in terms of ACC and NMI.

Future work will extend the proposed methods to spiking neural networks \cite{xiao2026spiking} to improve the computational efficiency. Furthermore, exploring the integration of large language models with UFS \cite{jeong2025llmselect} holds the potential to enhance feature selection capabilities on complex datasets.

\bibliographystyle{IEEEtran}
\bibliography{mybib}

@article{ray2021various,
  title={Various dimension reduction techniques for high dimensional data analysis: A review},
  author={Ray, Papia and Reddy, S Surender and Banerjee, Tuhina},
  journal={Artificial Intelligence Review},
  volume={54},
  number={5},
  pages={3473--3515},
  year={2021},
  publisher={Springer}
}

@article{erhard2020advances,
  title={Advances in high-dimensional quantum entanglement},
  author={Erhard, Manuel and Krenn, Mario and Zeilinger, Anton},
  journal={Nature Reviews Physics},
  volume={2},
  number={7},
  pages={365--381},
  year={2020},
  publisher={Nature Publishing Group UK London}
}

@article{li2024exploring,
  title={Exploring feature selection with limited labels: A comprehensive survey of semi-supervised and unsupervised approaches},
  author={Li, Guojie and Yu, Zhiwen and Yang, Kaixiang and Lin, Mianfen and Chen, CL Philip},
  journal={IEEE Transactions on Knowledge and Data Engineering},
  volume={36},
  number={11},
  pages={6124--6144},
  year={2024},
  publisher={IEEE}
}

@article{jiao2023survey,
  title={A survey on evolutionary multiobjective feature selection in classification: approaches, applications, and challenges},
  author={Jiao, Ruwang and Nguyen, Bach Hoai and Xue, Bing and Zhang, Mengjie},
  journal={IEEE Transactions on Evolutionary Computation},
  volume={28},
  number={4},
  pages={1156--1176},
  year={2024},
  publisher={IEEE}
}

@article{solorio2020review,
  title={A review of unsupervised feature selection methods},
  author={Solorio-Fern{\'a}ndez, Sa{\'u}l and Carrasco-Ochoa, J Ariel and Mart{\'\i}nez-Trinidad, Jos{\'e} Fco},
  journal={Artificial Intelligence Review},
  volume={53},
  number={2},
  pages={907--948},
  year={2020},
  publisher={Springer}
}

@article{theng2024feature,
  title={Feature selection techniques for machine learning: a survey of more than two decades of research},
  author={Theng, Dipti and Bhoyar, Kishor K},
  journal={Knowledge and Information Systems},
  volume={66},
  number={3},
  pages={1575--1637},
  year={2024},
  publisher={Springer}
}

@article{li2017feature,
  title={Feature selection: A data perspective},
  author={Li, Jundong and Cheng, Kewei and Wang, Suhang and Morstatter, Fred and Trevino, Robert P and Tang, Jiliang and Liu, Huan},
  journal={ACM computing surveys},
  volume={50},
  number={6},
  pages={1--45},
  year={2017},
  publisher={ACM New York, NY, USA}
}

@article{he2005laplacian,
  title={Laplacian score for feature selection},
  author={He, Xiaofei and Cai, Deng and Niyogi, Partha},
  journal={Advances in Neural Information Processing Systems},
  volume={18},
  year={2005}
}

@inproceedings{yang2011l21,
  title={$\ell_{2,1}$-norm regularized discriminative feature selection for unsupervised learning},
  author={Yang, Yi and Shen, Heng Tao and Ma, Zhigang and Huang, Zi and Zhou, Xiaofang},
  booktitle={IJCAI International Joint Conference on Artificial Intelligence},
  year={2011}
}

@inproceedings{nie2016unsupervised,
  title={Unsupervised feature selection with structured graph optimization},
  author={Nie, Feiping and Zhu, Wei and Li, Xuelong},
  booktitle={Proceedings of the AAAI Conference on Artificial Intelligence},
  volume={30},
  number={1},
  year={2016}
}

@article{liu2020robust,
  title={Robust neighborhood embedding for unsupervised feature selection},
  author={Liu, Yanfang and Ye, Dongyi and Li, Wenbin and Wang, Huihui and Gao, Yang},
  journal={Knowledge-Based Systems},
  volume={193},
  pages={105462},
  year={2020},
  publisher={Elsevier}
}

@article{li2015unsupervised,
  title={Unsupervised feature selection via nonnegative spectral analysis and redundancy control},
  author={Li, Zechao and Tang, Jinhui},
  journal={IEEE Transactions on Image Processing},
  volume={24},
  number={12},
  pages={5343--5355},
  year={2015},
  publisher={IEEE}
}

@article{zhou2024unsupervised,
  title={Unsupervised discriminative feature selection via contrastive graph learning},
  author={Zhou, Qian and Wang, Qianqian and Gao, Quanxue and Yang, Ming and Gao, Xinbo},
  journal={IEEE Transactions on Image Processing},
  volume={33},
  pages={972--986},
  year={2024},
  publisher={IEEE}
}

@article{li2026unsupervised,
  author  = {Li, Yan and Sun, Defeng and Zhang, Liping},
  title   = {Unsupervised Feature Selection via Nonnegative Orthogonal Constrained Regularized Minimization},
  journal = {Journal of Machine Learning Research},
  year    = {2026},
  volume  = {27},
  number  = {39},
  pages   = {1--44}
}

@article{greenacre2022principal,
  title={Principal component analysis},
  author={Greenacre, Michael and Groenen, Patrick JF and Hastie, Trevor and d’Enza, Alfonso Iodice and Markos, Angelos and Tuzhilina, Elena},
  journal={Nature Reviews Methods Primers},
  volume={2},
  number={1},
  pages={100},
  year={2022},
  publisher={Nature Publishing Group UK London}
}

@article{xiu2023sparsity,
  title={A sparsity-aware fault diagnosis framework focusing on accurate isolation},
  author={Xiu, Xianchao and Miao, Zhonghua and Liu, Wanquan},
  journal={IEEE Transactions on Industrial Informatics},
  volume={19},
  number={2},
  pages={1356--1365},
  year={2023},
  publisher={IEEE}
}

@article{zou2006sparse,
  title={Sparse principal component analysis},
  author={Zou, Hui and Hastie, Trevor and Tibshirani, Robert},
  journal={Journal of Computational and Graphical Statistics},
  volume={15},
  number={2},
  pages={265--286},
  year={2006},
  publisher={Taylor \& Francis}
}

@article{li2023sparse,
  title={Sparse {PCA} via $\ell_{2,p}$-Norm Regularization for Unsupervised Feature Selection},
  author={Li, Zhengxin and Nie, Feiping and Bian, Jintang and Wu, Danyang and Li, Xuelong},
  journal={IEEE Transactions on Pattern Analysis and Machine Intelligence},
  volume={45},
  number={4},
  pages={5322--5328},
  year={2023},
  publisher={IEEE}
}

@article{nie2023learning,
  title={Learning feature-sparse principal subspace},
  author={Nie, Feiping and Tian, Lai and Wang, Rong and Li, Xuelong},
  journal={IEEE Transactions on Pattern Analysis and Machine Intelligence},
  volume={45},
  number={4},
  pages={4858--4869},
  year={2023},
  publisher={IEEE}
}

@article{gao2024principal,
  title={Principal component analysis with fuzzy elastic net for feature selection},
  author={Gao, Yunlong and Wu, Qinting and Xu, Zhenghong and Cao, Chao and Pan, Jinyan and Shao, Guifang and Nie, Feiping and Zhu, Qingyuan},
  journal={IEEE Transactions on Fuzzy Systems},
  volume={32},
  number={12},
  pages={6878--6890},
  year={2024},
  publisher={IEEE}
}

@article{xiu2025bi,
  title={Bi-sparse unsupervised feature selection},
  author={Xiu, Xianchao and Huang, Chenyi and Shang, Pan and Liu, Wanquan},
  journal={IEEE Transactions on Image Processing},
  volume  = {34},
  pages   = {7407--7421},
  year    = {2025},
  publisher={IEEE}
}

@article{liu2026bi,
  title={Bi-level unsupervised feature selection with $\ell_{2,0}$-norm},
  author={Liu, Jingjing and Ju, Xiansen and Xiu, Xianchao and Liu, Wanquan},
  journal={Pattern Recognition},
  pages={114687},
  year={2026},
  publisher={Elsevier}
}

@article{zhang2023structured,
  title={Structured sparsity optimization with non-convex surrogates of $\ell_{2,0}$-norm: A unified algorithmic framework},
  author={Zhang, Xin and Zheng, Jiyuan and Wang, Dong and Tang, Guang and Zhou, Zhi and Lin, Zhouchen},
  journal={IEEE Transactions on Pattern Analysis and Machine Intelligence},
  volume={45},
  number={5},
  pages={6386--6402},
  year={2023},
  publisher={IEEE}
}

@article{mirzaei2020deep,
  title={Deep feature selection using a teacher-student network},
  author={Mirzaei, Ali and Pourahmadi, Vahid and Soltani, Mehran and Sheikhzadeh, Hamid},
  journal={Neurocomputing},
  volume={383},
  pages={396--408},
  year={2020},
  publisher={Elsevier}
}

@inproceedings{hassanieh2024selective,
  title={Selective deep autoencoder for unsupervised feature selection},
  author={Hassanieh, Wael and Chehade, Abdallah},
  booktitle={Proceedings of the AAAI Conference on Artificial Intelligence},
  volume={38},
  number={11},
  pages={12322--12330},
  year={2024}
}

@article{ling2025self,
  title={Self-labeling and self-knowledge distillation unsupervised feature selection},
  author={Ling, Yunzhi and Nie, Feiping and Yu, Weizhong and Li, Xuelong},
  journal={IEEE Transactions on Knowledge and Data Engineering},
  volume={37},
  number={7},
  pages={4270--4284},
  year={2025},
  publisher={IEEE}
}

@article{monga2021algorithm,
  title={Algorithm unrolling: Interpretable, efficient deep learning for signal and image processing},
  author={Monga, Vishal and Li, Yuelong and Eldar, Yonina C},
  journal={IEEE Signal Processing Magazine},
  volume={38},
  number={2},
  pages={18--44},
  year={2021},
  publisher={IEEE}
}

@article{shlezinger2023model,
  title={Model-based deep learning},
  author={Shlezinger, Nir and Eldar, Yonina C},
  journal={Foundations and Trends in Signal Processing},
  volume={17},
  number={4},
  pages={291--416},
  year={2023},
  publisher={Emerald Publishing Limited}
}

@article{guo2025usb,
  title={{USB-Net}: Unfolding split {B}regman method with multi-phase feature integration for compressive imaging},
  author={Guo, Zhen and Gan, Hongping},
  journal={IEEE Transactions on Image Processing},
  volume={34},
  pages={925--938},
  year={2025},
  publisher={IEEE}
}

@article{liu2025star,
  title={{STAR-Net}: An interpretable model-aided network for remote sensing image denoising},
  author={Liu, Jingjing and Jin, Jiashun and Xiu, Xianchao and Zhang, Jianhua and Liu, Wanquan},
  journal={Pattern Recognition},
  pages={112496},
  year={2025},
  publisher={Elsevier}
}

@article{deka2026comprehensive,
  title={Comprehensive review of deep unfolding techniques for next-generation wireless communication systems},
  author={Deka, Sukanya and Deka, Kuntal and Nguyen, Nhan Thanh and Sharma, Sanjeev and Bhatia, Vimal and Rajatheva, Nandana},
  journal={IEEE Internet of Things Journal},
  year={2026},
  volume={13},
  number={6},
  pages={10379--10406},
  publisher={IEEE}
}

@article{boumal2020introduction,
  title={An introduction to optimization on smooth manifolds},
  author={Boumal, Nicolas},
  journal={Cambridge University Press},
  year={2023}
}

@article{zhu2025sparse,
  title={Sparse tensor {CCA} via manifold optimization for multi-view learning},
  author={Zhu, Yanjiao and Liu, Wanquan and Xiu, Xianchao and Sun, Jianqin},
  journal={IEEE Transactions on Circuits and Systems for Video Technology},
  year={2026},
  volume={36},
  number={5},
  pages={6299--6313},
  publisher={IEEE}
}

@article{xu20101,
  title={$L_{1/2}$ regularization},
  author={Xu, Zongben and Zhang, Hai and Wang, Yao and Chang, XiangYu and Liang, Yong},
  journal={Science China Information Sciences},
  volume={53},
  number={6},
  pages={1159--1169},
  year={2010},
  publisher={Springer}
}

@inproceedings{gregor2010learning,
  title={Learning fast approximations of sparse coding},
  author={Gregor, Karol and LeCun, Yann},
  booktitle={Proceedings of the 27th International Conference on International Conference on Machine Learning},
  pages={399--406},
  year={2010}
}

@inproceedings{yang2016deep,
  title={Deep {ADMM-Net} for compressive sensing {MRI}},
  author={Yang, Yan and Sun, Jian and Li, Huibin and Xu, Zongben},
  booktitle={Advances in Neural Information Processing Systems},
  pages={10--18},
  year={2016}
}

@article{boyd2011distributed,
  title={Distributed optimization and statistical learning via the alternating direction method of multipliers},
  author={Boyd, Stephen and Parikh, Neal and Chu, Eric and Peleato, Borja and Eckstein, Jonathan},
  journal={Foundations and Trends{\textregistered} in Machine learning},
  volume={3},
  number={1},
  pages={1--122},
  year={2011},
  publisher={Emerald Publishing Limited}
}

@article{zhang2026top,
  title={Top-$k$ Feature Selection in Sparse Learning via Accelerated Coordinate Descent Method},
  author={Zhang, Han and Gu, Yannian and Nie, Feiping and Li, Xuelong},
  journal={IEEE Transactions on Pattern Analysis and Machine Intelligence},
  year={2026},
  volume={48},
  number={6},
  pages={6880-6896},
  publisher={IEEE}
}

@article{fan2026robust,
  title={Robust sparse phase retrieval: statistical guarantee, optimality theory and convergent algorithm},
  author={Fan, Jun and Yan, Ailing and Xiu, Xianchao and Liu, Wanquan},
  journal={Inverse Problems},
  volume={42},
  number={4},
  pages={045014},
  year={2026},
  publisher={IOP Publishing}
}

@article{wan2026manifold,
  title={Manifold Regularized Non-negative {PCA} with Robust $\ell_{2,p}$-Norm Enhancement},
  author={Wan, Minghua and Chen, Taotao and Tan, Hai and Tang, Mingwei and Yang, Guowei},
  journal={Pattern Recognition},
  pages={113195},
  year={2026},
  publisher={Elsevier}
}

@article{cui2023unrolling,
  title={Unrolling nonnegative matrix factorization with group sparsity for blind hyperspectral unmixing},
  author={Cui, Chunyang and Wang, Xinyu and Wang, Shaoyu and Zhang, Liangpei and Zhong, Yanfei},
  journal={IEEE Transactions on Geoscience and Remote Sensing},
  volume={61},
  pages={1--12},
  year={2023},
  publisher={IEEE}
}

@article{chen2022learning,
  title={Learning to optimize: A primer and a benchmark},
  author={Chen, Tianlong and Chen, Xiaohan and Chen, Wuyang and Heaton, Howard and Liu, Jialin and Wang, Zhangyang and Yin, Wotao},
  journal={Journal of Machine Learning Research},
  volume={23},
  number={189},
  pages={1--59},
  year={2022}
}

@article{shen2026causally,
  title={Causally-Aware Unsupervised Feature Selection Learning},
  author={Shen, Zongxin and Huang, Yanyong and Wang, Dongjie and Ma, Minbo and Lv, Fengmao and Li, Tianrui},
  journal={IEEE Transactions on Image Processing},
  year={2026},
  volume={35},
  number={},
  pages={1011--1024},
  publisher={IEEE}
}

@article{xiao2026spiking,
  title={Spiking meets attention: Efficient remote sensing image super-resolution with attention spiking neural networks},
  author={Xiao, Yi and Yuan, Qiangqiang and Jiang, Kui and Huang, Wenke and Zhang, Qiang and Zheng, Tingting and Lin, Chia-Wen and Zhang, Liangpei},
  journal={Advances in Neural Information Processing Systems},
  volume={38},
  pages={65240--65259},
  year={2026}
}

@article{joukovsky2024interpretable,
  title={Interpretable neural networks for video separation: Deep unfolding {RPCA} with foreground masking},
  author={Joukovsky, Boris and Eldar, Yonina C and Deligiannis, Nikos},
  journal={IEEE Transactions on Image Processing},
  volume={33},
  pages={108--122},
  year={2024},
  publisher={IEEE}
}

@article{wu2026rpcanet,
  author={Wu, Fengyi and Dai, Yimian and Zhang, Tianfang and Ding, Yixuan and Yang, Jian and Cheng, Ming-Ming and Peng, Zhenming},
  journal={IEEE Transactions on Pattern Analysis and Machine Intelligence},
  title={{RPCANet}$^{++}$: Deep Interpretable Robust {PCA} for Sparse Object Segmentation},
  year={2026},
  volume={48},
  number={10},
  pages={12420--12440},
  publisher={IEEE}
}

@article{zhou2025low,
  title={Low-rank and sparse representation meet deep unfolding: A new interpretable network for hyperspectral change detection},
  author={Zhou, Chengle and He, Zhi and Dong, Jian and Li, Yunfei and Ren, Jinchang and Plaza, Antonio},
  journal={IEEE Transactions on Geoscience and Remote Sensing},
  volume={63},
  pages={1--16},
  year={2025},
  publisher={IEEE}
}

@inproceedings{schechtman2023orthogonal,
  title={Orthogonal directions constrained gradient method: from non-linear equality constraints to {S}tiefel manifold},
  author={Schechtman, Sholom and Tiapkin, Daniil and Muehlebach, Michael and Moulines, Eric},
  booktitle={The Thirty Sixth Annual Conference on Learning Theory},
  pages={1228--1258},
  year={2023},
  organization={PMLR}
}

@article{zhu2025joint,
  title={Joint sparse subspace clustering via fast $\ell_{2,0}$-norm constrained optimization},
  author={Zhu, Yanjiao and Xiu, Xianchao and Liu, Wanquan and Yin, Chuancun},
  journal={Expert Systems with Applications},
  volume={265},
  pages={125845},
  year={2025},
  publisher={Elsevier}
}

@article{foucart2011hard,
  title={Hard thresholding pursuit: an algorithm for compressive sensing},
  author={Foucart, Simon},
  journal={SIAM Journal on Numerical Analysis},
  volume={49},
  number={6},
  pages={2543--2563},
  year={2011},
  publisher={SIAM}
}

@inproceedings{vinh2009information,
  title={Information theoretic measures for clusterings comparison: is a correction for chance necessary?},
  author={Vinh, Nguyen Xuan and Epps, Julien and Bailey, James},
  booktitle={Proceedings of the 26th Annual International Conference on Machine Learning},
  pages={1073--1080},
  year={2009}
}

@article{jeong2025llmselect,
title={{LLM}-Select: Feature Selection with Large Language Models},
author={Daniel P Jeong and Zachary Chase Lipton and Pradeep Kumar Ravikumar},
journal={Transactions on Machine Learning Research},
year={2025}
}

\end{document}